\documentclass{article}

\PassOptionsToPackage{numbers, compress}{natbib}
\usepackage[main, final]{neurips_2026}

\usepackage{silence}
\usepackage[utf8]{inputenc}
\usepackage[T1]{fontenc}
\usepackage{hyperref}
\usepackage{url}
\usepackage{booktabs}
\usepackage{multirow}
\usepackage{amsfonts}
\usepackage{amsmath}
\usepackage{amssymb}
\usepackage{nicefrac}
\usepackage{microtype}
\usepackage{graphicx}
\usepackage{xcolor}
\usepackage{multirow}
\usepackage{algorithm}
\usepackage{algorithmic}
\usepackage{subcaption}
\usepackage{bm}
\usepackage{enumitem}
\usepackage{mathtools}
\usepackage{amssymb}
\usepackage{xltabular}
\newcolumntype{Y}{>{\centering\arraybackslash}X}

\usepackage{tabularx}
\usepackage{caption}
\usepackage{array}
\newcommand{\E}{\mathbf{E}}
\newcommand{\W}{\mathbf{W}}
\newcommand{\x}{\mathbf{x}}

\newcommand{\q}{\mathbf{q}}

\newcommand{\R}{\mathbb{R}}
\newcommand{\softmax}{\operatorname{softmax}}

\title{CORTEG: Foundation Models Enable Cross-Modality Representation Transfer from Scalp to Intracranial Brain Recordings}

\author{
  Liuyin Yang\textsuperscript{1}\thanks{Equal contribution.}\textsuperscript{\, }\thanks{Correspondence to: \texttt{liuyin.yang@kuleuven.be} \&\, \texttt{marc.vanhulle@kuleuven.be}}\,\,\,,\,
  Qiang Sun\textsuperscript{1}\footnotemark[1]\,\,\,,\,
  Bob Van Dyck\textsuperscript{1},\,
  Eva Calvo Merino\textsuperscript{1},\,
  Marc M. Van Hulle\textsuperscript{1}\footnotemark[2] \\
  \textsuperscript{1}Laboratory for Neuro- \& Psychophysiology, Department of Neurosciences, KU Leuven \\[0.12cm]
  \,\,Code \& Visual: \textcolor{magenta}{\url{https://github.com/LiuyinYang1101/CORTEG/tree/main}}
}

\begin{document}

\maketitle

\begin{abstract}
Intracranial electrocorticography (ECoG) offers high–signal-to-noise access to cortical activity for brain–computer interfaces, yet limited per-patient data has led most prior work to rely on small, subject-specific decoders that neglect information shared across patients. We investigate whether large pretrained scalp-EEG foundation models (EEG FMs) can be adapted to ECoG, enabling cross-patient learning and competitive decoding performance while calibrating to a held-out patient in $10$--$30$ minutes on a single GPU. We introduce \textbf{CORTEG}, a cross-modality transfer framework that combines a pretrained EEG FM backbone, an electrode-aware KNNSoftFourier spatial adapter, a dual-stream tokenizer for low-frequency and high-gamma activity, and a leave-one-subject-out fine-tuning strategy. We evaluate CORTEG on two challenging regression tasks: public finger-trajectory regression ($n=9$) and private audio-envelope regression ($n=16$). CORTEG achieves state-of-the-art mean performance on both tasks, achieving the highest mean correlation among the compared methods on the public finger benchmark and larger, statistically significant gains on the audio task and in low-data-per-patient calibration. We further validate CORTEG on the public BrainTreebank stereo-EEG classification tasks, where it also performs strongly. Furthermore, CORTEG outperforms the compared intracranial-EEG foundation models across the continuous regression tasks. Feature analyses align with neurophysiology, and latent representations capture low-dimensional finger-movement structure. Together, these results provide systematic evidence that scalp-EEG pretraining can be repurposed across intracranial recording modalities and tasks, enabling data-efficient neural decoders that can adapt to new patients.

\end{abstract}

%=============================================================================
\section{Introduction}
\label{sec:intro}
%=============================================================================

Intracranial electrocorticography (ECoG) records the high-fidelity cortical activity essential for robust, high-performance brain-computer interfaces (BCIs)~\cite{benabid2019exoskeleton, lorach2023walking, metzger2022generalizable, littlejohn2025streaming}. Yet ECoG remains difficult to scale: recordings are invasive, usually limited to short clinical monitoring periods, and collected with electrode grids that vary substantially across patients in number, coverage, and location. Consequently, most prior ECoG decoders rely on classical machine-learning methods or compact neural networks trained independently for each patient, rather than population models that exploit shared structure across subjects. Transformer architectures trained from scratch on only tens of minutes of data per patient have also shown suboptimal performance~\cite{ferdinand2025ecog}. Although some public intracranial datasets and foundation models pretrained directly on intracranial recordings are now available~\cite{wang2024braintreebank, wang2023brainbert, chau2025population, zhang2023brant}, scalp EEG offers a substantially larger and faster-growing pretraining ecosystem. EEG is non-invasive and easier to acquire at scale, while public intracranial datasets remain considerably smaller~\cite{aristimunha2026eegdash}. At the same time, EEG foundation models (EEG FMs) are developing rapidly, with various pretraining strategies and model architectures~\cite{el2026reve, chien2211maeeg, yang2023biot, jiang2024large, wang2025cbramod, yang2026_steegformer}. This motivates us to investigate whether EEG FMs can transfer to continuous ECoG decoding and support efficient adaptation to new patients.

Three issues separate the two modalities. \textbf{Geometry:} EEG uses standardized montages (e.g., 10--20); ECoG grids cover patient-specific cortex with typically 16--256 electrodes. \textbf{Spectrum:} EEG predominantly reflects low-frequency cortical activity, while beta/gamma and higher-frequency bands are substantially reduced or harder to isolate at the scalp because volume conduction through the head attenuates and spatially smooths cortical potentials, and high-frequency scalp recordings are increasingly contaminated by extracranial sources~\cite{crone2001induced}. ECoG covers the high-gamma band on which motor decoding depends. \textbf{Data scarcity:} ECoG recordings are scarce and patient-specific, making it essential to leverage data from other subjects while adapting efficiently to new patients.

\begin{figure}[t]
\centering
\includegraphics[width=\textwidth]{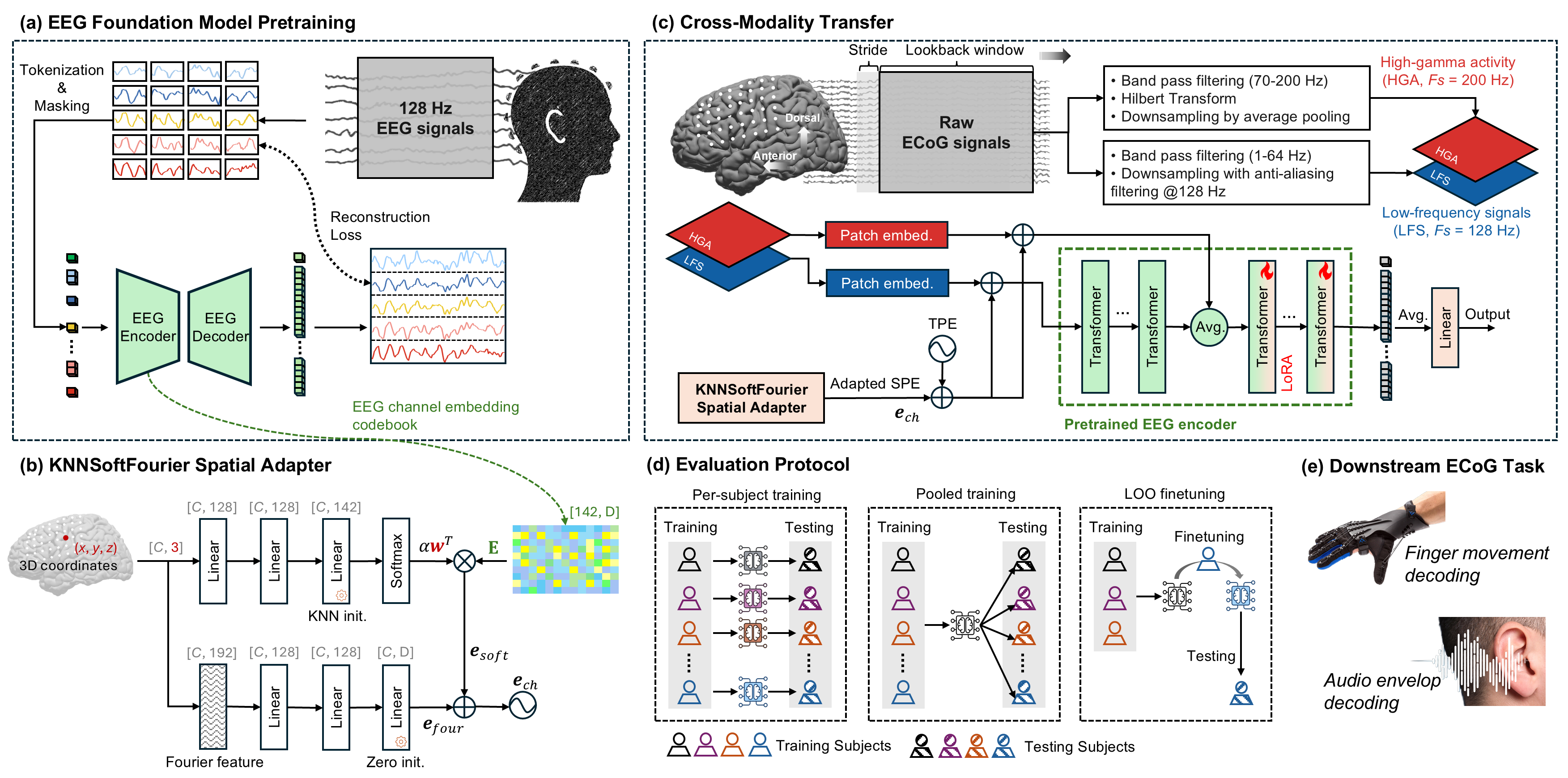}
\caption{
\textbf{Overview of CORTEG for cross-modality transfer from scalp EEG to ECoG.}
\textbf{(a)} A masked autoencoder is pretrained on 128~Hz scalp EEG, learning temporal representations and an EEG channel-embedding codebook $\E\in\R^{M'\times D}$.
\textbf{(b)} The KNNSoftFourier spatial adapter maps subject-specific ECoG electrode coordinates $\q=(x,y,z)$ into the pretrained EEG channel-embedding space using a soft codebook-interpolation branch and a Fourier residual branch.
\textbf{(c)} Raw ECoG is decomposed into low-frequency signals (LFS, 1--64~Hz, 128~Hz) and high-gamma activity (HGA, 70--200~Hz, 200~Hz), patch-embedded, combined with adapted spatial embeddings, and passed through the pretrained EEG encoder with LoRA on the final transformer blocks.
\textbf{(d)} Evaluation protocols include per-subject training, pooled training, and leave-one-subject-out fine-tuning (LOO-FT).
\textbf{(e)} Downstream tasks are continuous finger-trajectory decoding and audio-envelope decoding.
}
\label{fig:overview}
\vspace{-5.0pt}
\end{figure}

\paragraph{Contributions} We introduce \textbf{CORTEG} (Fig.~\ref{fig:overview}), a \underline{C}r\underline{O}ss-Modality \underline{R}epresentation \underline{T}ransfer framework that adapts a pretrained \underline{E}EG FM to ECo\underline{G}. Our main contributions are: 
\textbf{(1)} a parameter-efficient cross-subject deployment strategy, where LOO-FT with LoRA adapts a population model to a new patient in $10$--$30$ min while matching pooled and per-subject training; 
\textbf{(2)} task-dependent but consistent improvements from EEG pretraining across the evaluated ECoG decoding tasks;
and \textbf{(3)} the best mean correlation among compared methods on two continuous ECoG regression tasks, reaching the highest Pearson correlation $r{=}0.554$ for finger decoding and $r{=}0.339$ for audio-envelope decoding.

%=============================================================================
\section{Related Work}
\label{sec:related}
%=============================================================================

\textbf{Can scalp-EEG pretraining transfer cross-modally?} Self-supervised pretraining has produced a series of EEG FMs---EEGMAE~\cite{chien2211maeeg} (masked autoencoding on multi-channel EEG), BIOT~\cite{yang2023biot} (channel-agnostic architecture), LaBraM~\cite{jiang2024large} (neural tokenization with vector quantization), and CBraMod~\cite{wang2025cbramod} (channel-independent patching for heterogeneous montages). All of these models target scalp-EEG downstream tasks (e.g., sleep staging, emotion recognition, motor imagery, etc.). A recent pilot study found that EEG FMs can extract transferable low-frequency information from ECoG, but did not benchmark against state-of-the-art ECoG decoders~\cite{sun2026large}. In parallel, intracranial foundation models pretrained directly on iEEG have emerged, such as BrainBERT~\cite{wang2023brainbert}, the Population Transformer (PopT)~\cite{chau2025population}, and Brant~\cite{zhang2023brant}, which constitute a natural alternative to cross-modal scalp-EEG transfer. We benchmark CORTEG against all three in \S\ref{app:fm}.

\textbf{ECoG decoding} Finger trajectory regression from ECoG has been studied extensively~\cite{kubanek2009decoding, xie2018decoding, yao2022fast,faes2022single,lin2025towards}, with methods ranging from linear models on hand-crafted features to deep neural networks. Audio envelope decoding from cortical recordings has also attracted significant interest~\cite{vanthornhout2018speech,accou2023decoding,defossez2023decoding}, mainly as a simple readout of auditory stimulus tracking and as a first step toward richer acoustic or linguistic targets such as spectrograms, mel-frequency bands, or phonetic features~\cite{bellier2023music,evanson2025emergence}. These approaches typically train a separate model per subject on tens of minutes of recording. 

\textbf{Cross-subject and cross-modal transfer in neuroscience} Cross-subject transfer within-modality has been explored via domain adaptation, Euclidean alignment~\cite{he2019transfer}, and pooled multi-subject decoding with electrode projection~\cite{peterson2021generalized}. Cross-modal transfer between recording modalities has been studied for fMRI-to-EEG~\cite{huster2012methods} and MEG-to-EEG~\cite{henson2011parametric}, typically via source-space projection. Cross-modal transfer from scalp EEG to intracranial ECoG remains largely unexplored owing to the geometric and spectral mismatches discussed in \S\ref{sec:intro}.

\textbf{Parameter-efficient adaptation} LoRA~\cite{hu2021lora} and other adapter techniques~\cite{houlsby2019parameter} enable efficient finetuning of large pretrained models. We follow the standard practice to fine-tune our EEG FM.

\textbf{Spatial positional encoding} Our KNNSoftFourier adapter draws on Fourier feature positional encoding from neural radiance fields~\cite{mildenhall2021nerf,tancik2020fourier} and on Nadaraya--Watson kernel regression~\cite{nadaraya1964estimating,watson1964smooth}. We combine them to map coordinate spaces via a frozen pretrained EEG channel embedding codebook.

%=============================================================================
\section{Method}
\label{sec:method}
%=============================================================================

%-----------------------------------------------------------------------------
\subsection{Problem Formulation and Architecture}
\label{sec:method:problem}
%-----------------------------------------------------------------------------
\paragraph{EEG FM backbone choice} We use ST-EEGFormer~\cite{yang2026_steegformer} as our pretrained EEG FM backbone. Among current EEG FMs, it offers (a) a ViT-style~\cite{dosovitskiy2020image} architecture with an explicit, position-indexed channel-embedding codebook to which the spatial adapter (\S\ref{sec:method:adapter}) can attach to bridge the EEG/ECoG montage gap, and (b) strong downstream performance on EEG benchmarks~\cite{yang2026_steegformer}, including winning the NeurIPS~2025 EEG Foundation challenge regression task~\cite{aristimunha2025eeg}. Empirically, the choice matters: LaBraM, CBraMod, and the time-series FM MantisV2 evaluated as drop-in alternatives (\S\ref{sec:results:ablation}) all underperform ST-EEGFormer. ST-EEGFormer pretraining yields a ViT encoder backbone $f_\theta$ and a channel-embedding codebook $\E \in \R^{M' \times D}$ ($M'{=}142$, $D{=}512$ for the small model).
\vspace{-8pt}
\paragraph{Goal} To use per-subject ECoG $\{(\x_\text{lo}, \x_\text{hi})\}$ with $C_s$ subject-specific electrodes and regress to target $y \in \R^{d_\text{out}}$. This transfer requires (i) mapping subject ECoG coordinates $\q$ to EEG channel embedding space $\E$, (ii) bridging the scalp-to-cortical spectral gap, and (iii) generalizing across heterogeneous electrode layouts.
\vspace{-8pt}
\paragraph{Dual-stream tokenizer} The low-frequency stream $\x_\text{lo}$ is unfolded into $S$ temporal patches of length $p_\text{lo}$ and projected to dim $D$. The high-gamma stream $\x_\text{hi}$ uses a separate projection and patch size, pooled to the same number of tokens $S$. Both streams receive the same additive channel embedding $\mathbf{e}_\text{ch}^{(c)}$ (\S\ref{sec:method:adapter}) and a fixed sinusoidal temporal embedding. Low-stream tokens carry a learnable \texttt{[CLS]} token. The dual-stream design leverages the EEG FM's ability to capture low-frequency signals while augmenting them with high-gamma dynamics essential for ECoG decoding.
\vspace{-8pt}
\paragraph{Late fusion and LoRA adaptation} We use late fusion so that the low-frequency ECoG stream is first processed by the pretrained EEG backbone before being combined with high-gamma activity. This allows the early layers to extract EEG-like low-frequency representations, while the final layers focus on cross-band integration and task adaptation. The low stream traverses the first $L{-}K$ transformer layers (frozen), with $L$ the total number of transformer layers. At block $L{-}K$, the high-gamma tokens are averaged with the encoder-processed low-stream tokens, $\mathbf{Z}_\text{merged}{=}\tfrac{1}{2}(\mathbf{Z}_\text{lo}^{(L-K)}{+}\mathbf{Z}_\text{hi})$, and the merged sequence traverses the remaining $K$ layers jointly (default $K{=}4$). Final tokens are mean-pooled and projected via a linear head. We adapt those $K$ shared layers via LoRA~\cite{hu2021lora}. $\W' = \W + (\alpha/r)\mathbf{B}\mathbf{A}$ on each $\{\text{QKV},\text{O},\text{fc1},\text{fc2}\}$ projection ($r{=}4$, $\alpha{=}16$, $\mathbf{A}$ Kaiming-uniform, $\mathbf{B}{=}\mathbf{0}$). The base weights are frozen, so the trainable parameters are LoRA $\mathbf{A},\mathbf{B}$, the spatial adapter, and the regression head. The frozen early layers stay specialized for extracting low-frequency EEG-space representations, while cross-modal ECoG adaptation is restricted to the final layers after the low/high-gamma fusion.

%-----------------------------------------------------------------------------
\subsection{KNNSoftFourier Spatial Adapter}
\label{sec:method:adapter}
%-----------------------------------------------------------------------------

The spatial adapter $g_\phi: \R^3 \to \R^D$ produces channel embeddings $\mathbf{e}_\text{ch}(\q) = \mathbf{e}_\text{soft}(\q) + \mathbf{e}_\text{four}(\q)$ from a soft branch and a Fourier residual branch (Fig.~\ref{fig:overview} (b)); the soft-branch attention before and after training is further visualized in App.~Fig.~\ref{fig:adapter_attention}.
\vspace{-8pt}
\paragraph{Soft branch} A 3-layer MLP maps $\q$ to softmax attention weights $w_j(\q)$ over $M'$ EEG channels, giving $\mathbf{e}_\text{soft}(\q) = \alpha \sum_j w_j(\q) \E_j$ ($\alpha$ learnable, init $1$; $\E$ frozen, see Fig.~\ref{fig:overview} (b), top path). Only the bias of the last linear layer is initialized from the $k$-NN Gaussian kernel over EEG positions ($k{=}8$, $\sigma{=}$ median NN distance), so that $\softmax(\text{MLP}(\q))$ approximates the Nadaraya--Watson estimator~\cite{nadaraya1964estimating,watson1964smooth} at initialization; the upstream layers and last-layer weights are at default Kaiming init and are free to deform the kernel during training.
\vspace{-8pt}
\paragraph{Fourier residual} The same coordinate $\q$ is also routed through a positional-encoding branch (Fig.~\ref{fig:overview} (b), bottom path): $\gamma(\q){=}[\sin(f_i q_d), \cos(f_i q_d)]$ with $F{=}32$ log-spaced frequencies and three axes, then passed through a 3-layer MLP whose final projection is zero-init, so $\mathbf{e}_\text{four}{\equiv}\mathbf{0}$ at the start of training. The residual learns position-specific corrections from ECoG data that the soft branch's codebook-affine output cannot express. Branch roles (soft = task-relevant EEG attention; Fourier = position-specific differentiation) are validated in \S\ref{sec:results:ablation}.

%-----------------------------------------------------------------------------
\subsection{Leave-One-Subject-Out Fine-Tuning (LOO-FT)}
\label{sec:method:loo}
%-----------------------------------------------------------------------------
To adapt CORTEG to a new subject without retraining the population model, we use a two-stage LOO-FT strategy as shown in Fig.~\ref{fig:overview} (d). \textbf{Stage 1 (pooled training)}: for each held-out subject $s^*$, train CORTEG on the remaining $N{-}1$ subjects in a single pooled run with subject-interleaved batching. \textbf{Stage 2 (subject-specific calibration)}: load the Stage-1 checkpoint and fine-tune the spatial adapter, LoRA, and regression head on $s^*$'s training split with a fraction-aware adapter learning-rate boost ($10{\times}$ the base LR at recording fraction $f{\geq}0.25$, $2{\times}$ at $f{=}0.1$, see Fig.~\ref{fig:decoding}\,(d,e)). The large LR lets the adapter quickly escape the Stage-1 LOO bias on $s^*$'s held-out channel layout while keeping LoRA priors at the base LR. Full hyperparameters and failure-mode ablations are in App.~\ref{app:recipe}.

%=============================================================================
\section{Experimental Setup}
\label{sec:setup}
%=============================================================================

\paragraph{Datasets} We evaluate on two challenging ECoG regression tasks. \textbf{Finger trajectory regression} (public Stanford \textit{fingerflex} dataset ~\cite{miller2019library}): 9 subjects with $C \in [38, 64]$ ECoG channels performing individual finger flexion. The task aims to predict 5-finger positions ($d_\text{out}=5$) every 0.04 s with a 1-second lookback window of ECoG signals. \textbf{Audio envelope regression} (private dataset collected on patients during their epilepsy monitoring): 16 subjects with $C \in [16, 64]$ channels listening to continuous audio. The task aims to predict the broadband audio envelope ($d_\text{out}{=}1$) every $0.05$\,s with a $1$-second lookback window of ECoG signals. All subjects use MNI-coordinate electrode localization. For the cross-dataset foundation-model comparison (\S\ref{sec:results:btb}), we additionally evaluate on the public BrainTreebank stereo-EEG dataset~\cite{wang2024braintreebank}, which contains naturalistic language recordings collected during movie viewing. We use $10$ subjects and two binary classification tasks: a word-vs.-non-word benchmark and a sentence-initial vs.\ mid-sentence word task. Dataset and preprocessing details are provided in App.~\ref{app:data_prep}.

\paragraph{Splits and leakage prevention} For the public finger dataset, the train-val-test split is kept the same as previous works ~\cite{lin2025towards, sun2025spectro}, where the initial 400 seconds (or the first two-thirds) of the ECoG recordings and synchronized finger trajectories are used for training, with the remaining data reserved for testing. Within the training set, the last 10\% is used for validation. For the audio dataset, we split the continuous ECoG and audio recordings into 80\% for training, 10\% for validation, and reserve the last 10\% as the test set. For both datasets, the validation split is used for early stopping and hyperparameter search. Per-subject z-score normalization for both inputs and targets is computed on training windows only and applied identically at validation and test time. For the BrainTreebank tasks, we use chronological splits with a $7$ s gap between training and test events to ensure no overlap.

\paragraph{Baselines} We evaluate the following ECoG decoders. \textbf{Linear models}: ridge regression on the low-frequency and high-gamma streams. \textbf{Classic deep learning models trained per subject}: CNN-LSTM ~\cite{lin2025towards}, HiLoFuseNet~\cite{sun2025spectro} (depthwise-separable conv + LSTM on dual-stream), LSTM variants ~\cite{sun2025spectro} (LSTM\_LFS on raw 128~Hz, LSTM\_HGA on high-gamma 200~Hz), DeepFingerNet~\cite{tao2025deepfingernet} (nested U-Net with Morlet wavelet inputs), and HOPLS (tensor PLS). \textbf{Other EEG/Time-series FM models} (ablation study): LaBraM~\cite{jiang2024large}, CBraMod~\cite{wang2025cbramod}, MantisV2~\cite{feofanov2026mantisv2}, and REVE~\cite{el2026reve} adapted to our framework with their pretrained encoders. \textbf{Intracranial-EEG foundation models}: BrainBERT~\cite{wang2023brainbert}, the Population Transformer (PopT)~\cite{chau2025population}, and Brant~\cite{zhang2023brant}, evaluated under multiple adaptation strategies as direct intracranial-pretraining alternatives to CORTEG (\S\ref{app:fm}). \textbf{CORTEG variants}: pooled CORTEG, per-subject CORTEG, and LOO-FT CORTEG as defined in \S\ref{sec:method:loo}. Moreover, we also scale CORTEG by increasing the backbone from Small (default, $25.6$M) to Base ($85.6$M) and further to Large ($303$M) ST-EEGFormer.

\paragraph{Implementation} CORTEG uses the ST-EEGFormer small variant ($D{=}512$, $L{=}8$) and trained with AdamW ($\beta_1{=}0.9, \beta_2{=}0.999$), bfloat16 mixed precision. Pooled CORTEG: batch size 64, learning rate $3{\times}10^{-3}$ (audio $10^{-3}$), cosine schedule with 10-epoch warmup, weight decay $0.005$, 100-epoch budget with patience 90 on mean (audio 50) per-subject $\bar{r}$. LoRA rank $r{=}4$, $\alpha_\text{LoRA}{=}16$, applied to the last 4 transformer blocks across $\{\text{QKV},\text{O},\text{fc1},\text{fc2}\}$. KNNSoftFourier: $k{=}8$ neighbors, $F{=}32$ Fourier frequencies. Compute: NVIDIA RTX 5090 (local) plus occasional H100 (HPC) for scaling sweeps. A pooled CORTEG-S run takes approximately $2$ GPU-hours on a single RTX 5090 ($\approx5$\,h for Base and $\approx13$\,h for Large); LOO-FT Stage-2 takes 10--30 minutes per held-out subject.

%=============================================================================
\section{Results}
\label{sec:results}
%=============================================================================

%-----------------------------------------------------------------------------
\subsection{Comparison to Per-Subject Baselines}
\label{sec:results:baselines}
%-----------------------------------------------------------------------------

\begin{table}[t]
\caption{Comparison with classic baselines and iEEG FMs. Mean Pearson $r$ ($\pm$ cross-subject SD). \textcolor{blue}{\textbf{Best in bold blue}} and \textbf{second-best in bold black}. $^{\dagger}$From~\cite{sun2025spectro}; $^{\ddagger}$from~\cite{tao2025deepfingernet}; all other results use the shared protocol in \S\ref{sec:setup}. $^{\diamond}$SD omitted for continuously evaluated Ridge audio. $^{\S}$Best iEEG-FM adaptations in Table~\ref{tab:fm} (App.~\ref{app:fm}).}
\label{tab:main}
\centering
\footnotesize
\setlength{\tabcolsep}{3pt}
\begin{tabular}{@{}lcccl@{}}
\toprule
\textbf{Method} & \textbf{Trainable Parameters} & \textbf{Finger ($r$, $n{=}9$)} & \textbf{Audio ($r$, $n{=}16$)} & \textbf{Type} \\
\midrule
Ridge\_LFS              & ---       & $0.181 {\pm} 0.085$                          & $0.224^{\diamond}$              & Linear, LFS only \\
Ridge\_HGA              & ---       & $0.336 {\pm} 0.095$                          & $0.175^{\diamond}$              & Linear, HGA only \\
PLS                     & ---       & $0.402^{\dagger} {\pm} 0.115$                & $0.203 {\pm} 0.151$             & Linear, multi-band wavelet \\
HOPLS                   & ---       & $0.364^{\dagger} {\pm} 0.123$                & $0.203 {\pm} 0.142$             & Tensor latent-factor, multi-band \\
LSTM\_LFS               & 361\,K    & $0.276 {\pm} 0.164$                          & $0.198 {\pm} 0.162$             & Deep, LFS only \\
LSTM\_HGA               & 361\,K    & $0.485^{\dagger} {\pm} 0.136$                & $0.137 {\pm} 0.167$             & Deep, HGA only \\
CNN-LSTM                & 557\,K    & $0.411^{\dagger} {\pm} 0.124$                & $0.261 {\pm} 0.195$             & Deep, multi-band wavelet \\
DeepFingerNet           & 1.16\,M   & $0.542^{\ddagger} {\pm} 0.129$               & $0.085 {\pm} 0.144$             & Deep, multi-band wavelet (UNet++) \\
HiLoFuseNet             & 334\,K    & $0.534^{\dagger} {\pm} 0.138$                & $0.259 {\pm} 0.203$             & Deep, LFS\,+\,HGA dual-stream \\
\midrule
BrainBERT               & varies & $0.053 {\pm} 0.056$ & $0.057 {\pm} 0.066$ & iEEG FM$^{\S}$ \\
PopT                    & varies & $0.063 {\pm} 0.046$ & $0.039 {\pm} 0.064$ & iEEG FM$^{\S}$ \\
Brant                   & varies & $0.028 {\pm} 0.031$ & $0.022 {\pm} 0.035$ & iEEG FM$^{\S}$ \\
\midrule
CORTEG (ours)           & 297\,K    & $0.539 {\pm} 0.140$                          & $0.250 {\pm} 0.226$             & Per-subject \\
CORTEG (ours)           & 297\,K    & $\mathbf{0.551 {\pm} 0.147}$                 & $\mathbf{0.331 {\pm} 0.184}$    & LOO-FT \\
\textbf{CORTEG (ours)}  & 297\,K    & \textcolor{blue}{$\mathbf{0.554 {\pm} 0.154}$} & \textcolor{blue}{$\mathbf{0.339 {\pm} 0.170}$} & Pooled \\
\bottomrule
\end{tabular}
\vspace{-4.0pt}
\end{table}

Table~\ref{tab:main} summarizes the main decoding results. Pooled CORTEG achieves the highest mean Pearson $r$ on both tasks. On the public finger benchmark, it reaches $r{=}0.554$, compared with $0.542$ for DeepFingerNet and $0.534$ for HiLoFuseNet. These differences are modest and not statistically significant across the $n{=}9$ subjects, indicating that CORTEG is statistically comparable to the strongest task-specific decoders. Under larger-scale training, CORTEG reaches $r{=}0.583$ (Base) and $0.577$ (Small). The corresponding statistics are reported in App.~\ref{app:decoding_protocols}, Table~\ref{tab:hpc_stats}. On the audio task, the improvement is larger. Pooled CORTEG reaches $r{=}0.339$, compared with $0.261$ for CNN-LSTM and $0.259$ for HiLoFuseNet, and the differences remain significant after Bonferroni correction ($p{<}0.01$; Fig.~\ref{fig:decoding} (a)). LOO-FT remains close to pooled training on both tasks ($0.551$ finger; $0.331$ audio), showing that a population model can be adapted efficiently to new patients without retraining a separate model from scratch. Full per-subject and per-finger results are provided in App.~\ref{app:per_subject_results}. Interestingly, the three intracranial FMs perform substantially worse on both regression tasks. Full adaptation results are provided in App.~\ref{app:fm}.

%-----------------------------------------------------------------------------
\subsection{Decoding Results Across Both Tasks}
\label{sec:results:decoding}
%-----------------------------------------------------------------------------

\begin{figure}[t]
\centering
\includegraphics[width=\textwidth]{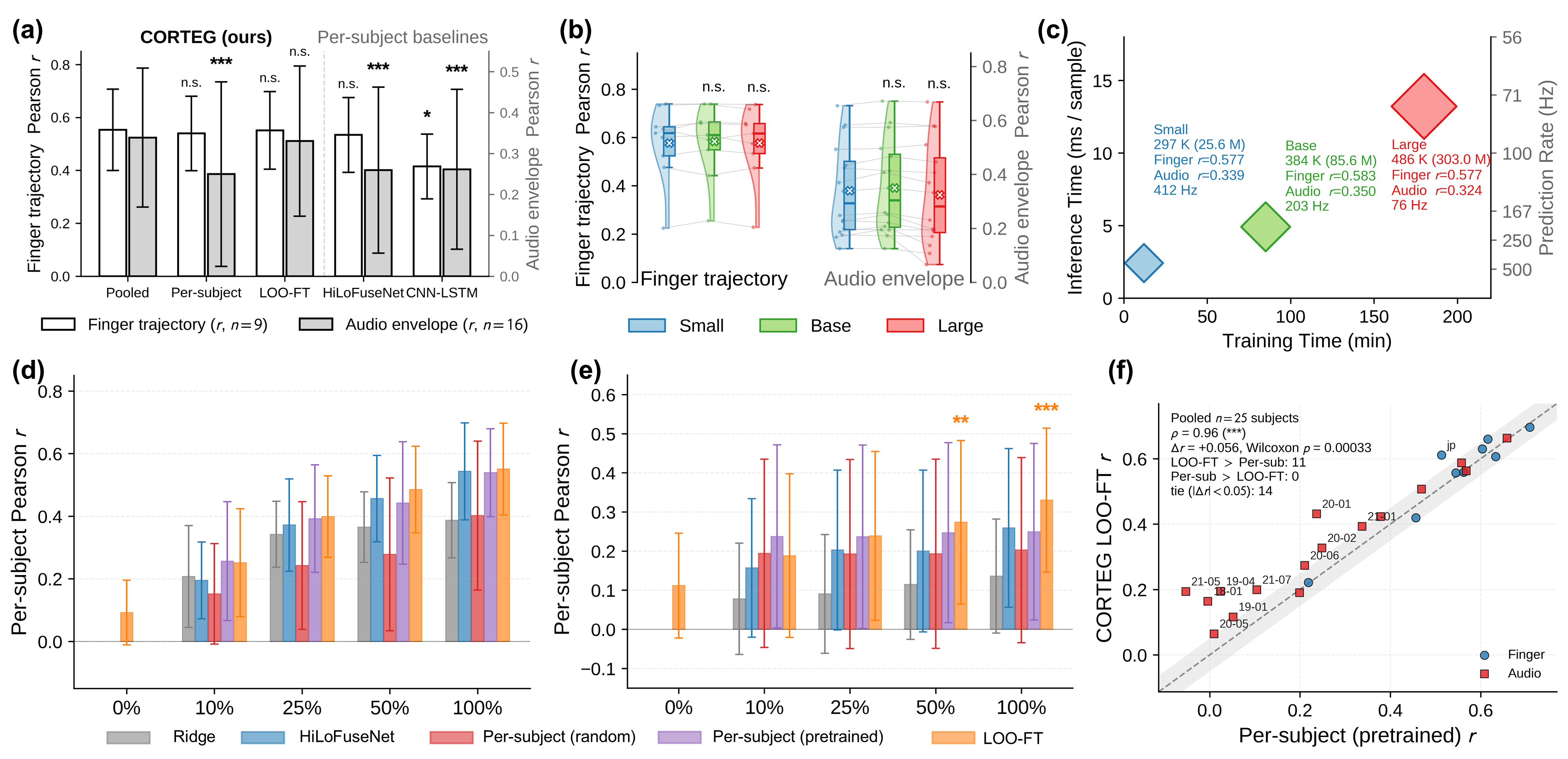}
\caption{\textbf{CORTEG decoding results.}
\textbf{(a)} Full-data comparison on finger ($n{=}9$) and audio ($n{=}16$) decoding; bars show mean $\pm$ SD, with paired Wilcoxon tests against pooled CORTEG (Bonferroni corrected).
\textbf{(b)} Per-subject performance of CORTEG-Small/Base/Large.
\textbf{(c)} Training and inference cost across backbone sizes; marker area indicates total parameter count.
\textbf{(d,e)} Performance with $0\%$, $10\%$, $25\%$, $50\%$, and $100\%$ of patient-specific calibration data on finger (d) and audio (e); $0\%$ denotes zero-shot decoding.
\textbf{(f)} Full-data per-subject comparison between pretrained per-subject CORTEG and LOO-FT; the shaded region indicates differences within $|\Delta r|{<}0.05$.}
\vspace{-10pt}
\label{fig:decoding}
\end{figure}

Figure~\ref{fig:decoding} shows model performance comparisons on full-data training (a), backbone scaling (b), compute-performance trade-off (c), and low-data analyses (d,e) with a full-data per-subject comparison (f). Corresponding experimental protocols are documented in App.~\ref{app:decoding_protocols}.

\textbf{LOO-FT reaches pooled-level performance} Figure~\ref{fig:decoding} (a) shows similar performance for LOO-FT and pooled CORTEG on both tasks, with no significant differences (paired Wilcoxon: $p{=}0.65$ for finger; $p{=}0.82$ for audio). Figure~\ref{fig:decoding} (f) further compares LOO-FT with per-subject CORTEG. Their subject-wise performance is strongly correlated (Spearman $\rho{=}0.96$), while LOO-FT improves by more than $|\Delta r|{=}0.05$ for $11/25$ subjects, with no subject showing a larger decrease. These results show that a population model can be efficiently adapted to a new patient through Stage-2 fine-tuning while retaining performance comparable to pooled and per-subject training.

\textbf{Backbone scaling and compute trade-off} Figure~\ref{fig:decoding} (b,c) evaluates CORTEG-Small/Base/Large under a larger-scale training sweep, using the best merge point $K$ for each backbone from Table~\ref{tab:merge}. CORTEG-Base achieves the highest mean performance, but differences across backbone sizes are not statistically significant. In contrast, inference time increases from $2.4$ to $13.2$ ms/sample on an RTX 5090 in fp32. CORTEG-Small therefore provides the most favorable performance--efficiency trade-off while supporting real-time inference. These results use a larger-scale training protocol than Table~\ref{tab:main}. Details are provided in App.~\ref{app:decoding_protocols}.

\textbf{Learning from limited data} Figure~\ref{fig:decoding} (d,e) evaluates how different methods perform as the amount of patient-specific calibration data increases. At $0\%$, the Stage-1 population model already achieves positive mean correlations on both tasks (finger $r{=}0.092$, audio $r{=}0.112$), with most subjects above zero ($8/9$ finger, $14/16$ audio; App.~Table~\ref{tab:zeroshot}), although the performance remains limited. As calibration data increases, LOO-FT improves consistently and is generally competitive with or better than per-subject CORTEG, while outperforming randomly initialized CORTEG.

%-----------------------------------------------------------------------------
\subsection{Model Interpretation and Representation Analysis}
\label{sec:results:manifold}
%-----------------------------------------------------------------------------

\begin{figure}[t]
\centering
\includegraphics[width=0.9\textwidth]{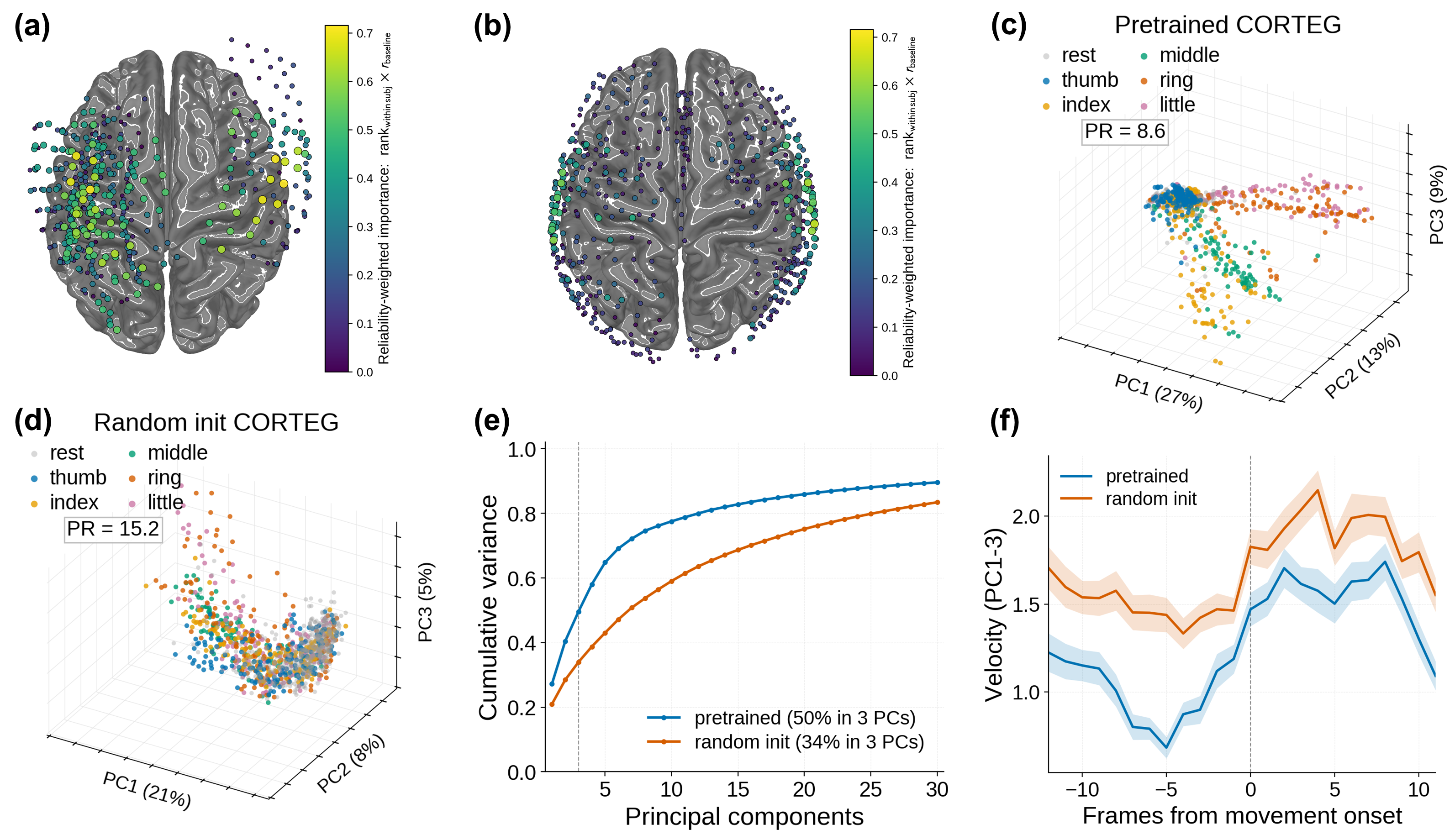}
\vspace{-6pt}
\caption{\textbf{Electrode importance and latent representations of CORTEG.}
\textbf{(a,b)} Per-electrode importance for finger trajectory (a) and audio-envelope (b) decoding, showing distinct sensorimotor and superior-temporal spatial patterns.
\textbf{(c,d)} 3D PCA of window-level CORTEG representations for finger-task subject S1 (\texttt{bp}), colored by dominant finger and rest (grey), for pretrained (c) and randomly initialized (d) CORTEG. The participation ratios are $8.6$ and $15.2$, respectively.
\textbf{(e)} Cumulative explained variance across principal components for pretrained and randomly initialized CORTEG.
\textbf{(f)} Representational velocity $\|\Delta z_{1:3}\|$ in the first three PCs around movement onset ($n{=}184$ events; $t{=}0$ denotes onset). Details are provided in App.~\ref{app:manifold_protocols}.}
\label{fig:neuroscience}
\end{figure}

\textbf{Channel importance} Fig.~\ref{fig:neuroscience} (a,b) measures per-electrode importance by the decrease in $r$ after removing each channel and projects the results onto one cortex template (top view). The two tasks show distinct spatial patterns, with finger-decoding importance concentrated around the sensorimotor cortex, whereas audio-decoding importance is concentrated around superior temporal regions. These patterns emerge without explicit task-specific anatomical supervision, suggesting that the model relies on localized, task-relevant cortical regions rather than diffuse contributions across the full grid. A full version of the same analysis, pooled across subjects on lateral and dorsal views, is shown in Fig.~\ref{fig:channel_importance}.

\textbf{Latent representation geometry.} Fig.~\ref{fig:neuroscience} (c--f) compares pretrained and randomly initialized CORTEG under the same architecture and training protocol for subject \texttt{bp}. Pretraining yields a lower-dimensional latent space, with participation ratio decreasing from $15.2$ to $8.6$, and concentrates more variance in the leading principal components (panel e). It also shows a clearer finger-dependent structure in the PCA projection and a sharper representational change around movement onset (panel f). These findings suggest that pretraining provides not just better features but a more compact, task-aligned subspace. Alternative LDA and dPCA projections show similar patterns (App.~\ref{app:projections}). Together, they suggest that EEG pretraining helps the model learn a more compact and task-structured latent representation.

%-----------------------------------------------------------------------------
\subsection{Ablation Study}
\label{sec:results:ablation}
%-----------------------------------------------------------------------------

\newsavebox{\ablationbox}
\newsavebox{\btbbox}
\newsavebox{\gatebox}
\newlength{\righttablegap}

\begin{table*}[t]
\centering

\begin{lrbox}{\ablationbox}
\begin{minipage}[t]{0.54\textwidth}
\vspace{0pt}
\centering

\captionof{table}{
Model ablation results. Each row modifies one component of pooled CORTEG-S.
Values are the mean Pearson $r$ across subjects; $\Delta$ is relative to the full model.
$^{\star}$Random initialization uses full fine-tuning.
$^{\P}$REVE uses its native coordinate encoding.
$^{\mathsection}$The broadband control is compared with its matched dual-stream run.
Full details are provided in App.~\ref{app:ablation}.
}
\label{tab:ablation}

\footnotesize
\setlength{\tabcolsep}{4pt}
\renewcommand{\arraystretch}{0.96}

\begin{tabular*}{\linewidth}{@{\extracolsep{\fill}}lcc@{}}
\toprule
\textbf{Configuration}
& \textbf{Finger $r$ ($\Delta$)}
& \textbf{Audio $r$ ($\Delta$)} \\
\midrule

\textbf{Full CORTEG-S}
& $\mathbf{0.554}$
& $\mathbf{0.339}$ \\

\addlinespace[2pt]
\multicolumn{3}{@{}l}{\textcolor{gray}{\textit{Foundation model}}} \\[-2pt]
LaBraM~\cite{jiang2024large}
& $0.372\,(-0.182)$
& $0.228\,(-0.111)$ \\
CBraMod~\cite{wang2025cbramod}
& $0.312\,(-0.242)$
& $0.045\,(-0.294)$ \\
MantisV2~\cite{feofanov2026mantisv2}
& $0.198\,(-0.356)$
& --- \\
REVE~\cite{el2026reve}$^{\P}$
& $0.529\,(-0.025)$
& $0.293\,(-0.046)$ \\
Random init$^{\star}$
& $0.510\,(-0.044)$
& $0.156\,(-0.183)$ \\

\addlinespace[2pt]
\multicolumn{3}{@{}l}{\textcolor{gray}{\textit{Adaptation strategy}}} \\[-2pt]
Full FT + adapter
& $0.542\,(-0.012)$
& $0.265\,(-0.074)$ \\
LoRA, no adapter
& $0.529\,(-0.025)$
& $0.299\,(-0.040)$ \\

\addlinespace[2pt]
\multicolumn{3}{@{}l}{\textcolor{gray}{\textit{Training strategy}}} \\[-2pt]
Per-subject
& $0.539\,(-0.015)$
& $0.250\,(-0.089)$ \\
LOO-FT
& $0.551\,(-0.003)$
& $0.331\,(-0.008)$ \\

\addlinespace[2pt]
\multicolumn{3}{@{}l}{\textcolor{gray}{\textit{Input stream}}} \\[-2pt]
Single broadband$^{\mathsection}$
& $0.534\,(-0.011)$
& --- \\
High-gamma only
& $0.420\,(-0.134)$
& $0.134\,(-0.205)$ \\
Low-frequency only
& $0.364\,(-0.190)$
& $0.273\,(-0.066)$ \\

\addlinespace[2pt]
\multicolumn{3}{@{}l}{\textcolor{gray}{\textit{Electrode geometry}}} \\[-2pt]
Random XYZ
& $0.479\,(-0.075)$
& $0.328\,(-0.011)$ \\
Zero XYZ
& $0.236\,(-0.318)$
& $0.153\,(-0.186)$ \\

\addlinespace[2pt]
\multicolumn{3}{@{}l}{\textcolor{gray}{\textit{Adapter branch}}} \\[-2pt]
Fourier-only
& $0.515\,(-0.039)$
& $0.330\,(-0.009)$ \\
Soft-only
& $0.443\,(-0.111)$
& $0.228\,(-0.111)$ \\

\bottomrule[\heavyrulewidth]
\end{tabular*}

\end{minipage}
\end{lrbox}

% ============================================================
% TABLE 3
% ============================================================
\begin{lrbox}{\btbbox}
\begin{minipage}[t]{0.43\textwidth}
\vspace{0pt}
\centering

\captionof{table}{
Generalization to BrainTreebank. Mean AUROC ($\pm$ cross-subject SD) across
$10$ subjects under the leakage-free protocol.
Task~A: sentence-initial vs.\ mid-sentence;
Task~B: word vs.\ non-word.
Each iEEG FM is shown in its best-evaluated configuration.
The CORTEG row uses layer-wise gated fusion.
Full results are provided in App.~\ref{app:fm}. $^{\dagger}$~Single-best-electrode arm (see Table~\ref{tab:btb_persubj}).
}
\label{tab:btb_main}

\scriptsize
\setlength{\tabcolsep}{2.5pt}
\renewcommand{\arraystretch}{1.04}

\begin{tabularx}{\linewidth}{
@{}
>{\raggedright\arraybackslash}X
c
c
@{}
}
\toprule
\textbf{Method} & \textbf{Task A} & \textbf{Task B} \\
\midrule

BrainBERT~\cite{wang2023brainbert}$^{\dagger}$
& $\mathbf{0.602{\pm}0.045}$
& $0.698{\pm}0.094$ \\

PopT~\cite{chau2025population}
& $0.600{\pm}0.084$
& \textcolor{blue}{$\mathbf{0.779{\pm}0.099}$} \\

Brant~\cite{zhang2023brant}$^{\dagger}$
& $0.587{\pm}0.041$
& $0.577{\pm}0.037$ \\

HiLoFuseNet
& $0.533{\pm}0.022$
& $0.721{\pm}0.089$ \\

\midrule

CORTEG, random init
& $0.538{\pm}0.054$
& $0.588{\pm}0.102$ \\

\textbf{CORTEG (ours)}
& \textcolor{blue}{$\mathbf{0.638{\pm}0.078}$}
& $\mathbf{0.749{\pm}0.134}$ \\

\bottomrule[\heavyrulewidth]
\end{tabularx}

\end{minipage}
\end{lrbox}

% ============================================================
% TABLE 4
% ============================================================
\begin{lrbox}{\gatebox}
\begin{minipage}[t]{0.43\textwidth}
\vspace{0pt}
\centering

\captionof{table}{
Layer-wise gated fusion compared with the default mean-pool fusion. Values show the change in Pearson $r$ ($\Delta r$). Bold values remain significant after correction. Full results are provided in App.~\ref{app:gate}.
}
\label{tab:gate_main}

\footnotesize
\setlength{\tabcolsep}{4pt}
\renewcommand{\arraystretch}{1.05}

\begin{tabularx}{\linewidth}{
@{}
>{\raggedright\arraybackslash}X
c
c
@{}
}
\toprule
\textbf{Scale}
& \textbf{Finger $\Delta r$}
& \textbf{Audio $\Delta r$} \\
\midrule

Small
& $+0.002$
& $\mathbf{-0.015}$ \\

Base
& $\mathbf{+0.018}$
& $\mathbf{+0.044}$ \\

Large
& $+0.010$
& $\mathbf{+0.129}$ \\

\bottomrule[\heavyrulewidth]
\end{tabularx}

\end{minipage}
\end{lrbox}

% ============================================================
% AUTOMATICALLY CALCULATE EXACT GAP
% ============================================================
\setlength{\righttablegap}{%
  \dimexpr
  \ht\ablationbox+\dp\ablationbox
  -\ht\btbbox-\dp\btbbox
  -\ht\gatebox-\dp\gatebox
  \relax
}

% ============================================================
% RENDER
% ============================================================
\noindent
\usebox{\ablationbox}%
\hfill
\begin{minipage}[t]{0.43\textwidth}
\vspace{0pt}

% Top-right table
\usebox{\btbbox}

% Exact automatically calculated space
\vspace*{\righttablegap}

% Bottom-right table
\usebox{\gatebox}

\end{minipage}

\end{table*}

Table~\ref{tab:ablation} evaluates the main components of CORTEG.

\textbf{(i) Pretraining and backbone choice} Removing pretraining reduces performance, particularly on audio ($\Delta r{=}-0.183$). Transfer quality also depends on the pretrained backbone. LaBraM, CBraMod, and MantisV2 perform substantially worse, whereas replacing ST-EEGFormer with REVE retains most of the performance. Thus, EEG pretraining is beneficial, and the choice of backbone remains important. This suggests that continuous ECoG decoding can serve as a challenging cross-modality benchmark for EEG foundation models, where better EEG representations may translate to intracranial signals and reflect more general neural signatures.

\textbf{(ii) Geometry and adaptation} Using the real electrode coordinates consistently outperforms random or zero coordinates, with the largest effect on finger decoding. The full KNNSoftFourier adapter also performs best overall. LoRA slightly outperforms full fine-tuning, indicating that parameter-efficient adaptation is sufficient for transferring the pretrained backbone.

\textbf{(iii) Training strategy} Pooled training gives the highest mean performance, while LOO-FT remains within $\Delta r{<}0.01$ on both tasks. Per-subject training performs similarly on finger but drops more clearly on audio. These results support LOO-FT as an effective method for adapting a population model to unseen patients.

\textbf{(iv) Frequency-stream contributions} The band contribution is task-dependent. High-gamma contributes more to finger decoding, whereas low-frequency activity is more useful for audio decoding. The full dual-stream model achieves the best overall performance, suggesting that combining these frequency components helps the model exploit task-relevant information from both bands. Additional stream, merge-point, and fusion analyses are provided in the Appendix.

\textbf{(v) Layer-wise gated fusion} We also test a lightweight layer-wise gated fusion that avoids searching for the best merge point $K$ and adds only ${\sim}17$K parameters. It improves both tasks at the Base and Large scales, while the results are mixed at the Small scale (Table~\ref{tab:gate_main}). Full results are provided in App.~\ref{app:gate}.
\subsection{Generalization Beyond ECoG Regression Tasks}
\label{sec:results:btb}

Additionally, we evaluate CORTEG and the three iEEG FMs on BrainTreebank~\cite{wang2024braintreebank}, a public sEEG dataset with two binary language-classification tasks. To test CORTEG's generalization, the setting differs from our main experiments in dataset, recording modality, and task type. Details about this task are provided in App.~\ref{app:btb}.

Unlike regression tasks, the iEEG FMs achieve substantially stronger performance on BrainTreebank (Table~\ref{tab:btb_main}). PopT reaches AUROC $0.779$ on the word-vs.-non-word task, while CORTEG remains competitive at $0.749$ and achieves the highest performance on sentence-position classification. EEG pretraining also improves CORTEG over the randomly initialized backbone. These results show that CORTEG generalizes beyond ECoG regression tasks. Meanwhile, the stronger performance of the iEEG FMs on BrainTreebank suggests that their poor results on our regression tasks reflect limited transfer to continuous regression rather than an implementation issue.

%=============================================================================
\section{Discussion}
\label{sec:discussion}
%=============================================================================

\paragraph{Is EEG FM to ECoG decoding feasible?}
To our knowledge, CORTEG provides one of the first systematic studies of transferring a scalp-EEG FM to intracranial ECoG decoding under matched per-subject baselines. Despite the gap in modality (extracranial vs.\ intracranial), sampling rate, montage, and frequency content, an EEG FM backbone trained on 142-channel scalp EEG, combined with a lightweight spatial adapter (KNNSoftFourier), matches or exceeds task-specific deep baselines on both finger trajectory ($r{=}0.554$) and audio envelope ($r{=}0.339$) decoding. The contribution of EEG pretraining is task-dependent: removing it causes a larger drop on audio ($\Delta r{=}{-}0.183$) than on finger decoding ($\Delta r{=}{-}0.044$). The stream ablations provide a possible explanation: high-gamma contributes more to finger decoding, whereas low-frequency activity is more informative for audio, a pattern that also holds with the REVE backbone. This is consistent with stronger transfer when the task-relevant information overlaps more closely with the frequency range represented in scalp EEG pretraining. The backbone model choice also matters. REVE retains most of CORTEG's performance, whereas LaBraM, CBraMod, and MantisV2 transfer less effectively. Future work should investigate whether stronger EEG backbones and model-specific adaptation strategies can further improve cross-modal transfer.

\paragraph{Why not start from an intracranial FM?}
A natural alternative is to adapt a foundation model pretrained directly on intracranial recordings. We benchmark BrainBERT, PopT, and Brant under multiple adaptation strategies and find that all remain weak on continuous ECoG regression. Importantly, model-specific optimizations improve these baselines. For example, BrainBERT and PopT benefit from better electrode-level readout on some subjects. Temporal decoding also improves most configurations, whereas last-$N$ fine-tuning does not consistently help. Although further gains may be possible with more tailored adaptations, our results suggest that existing iEEG FMs do not easily transfer to continuous regression tasks. One possible explanation is that the representations learned by these models are less suited to the task-specific spatial patterns across electrodes and the temporally precise representations required for continuous regression. Our representation analysis also shows limited accessibility of high-gamma information in their frozen embeddings. Overall, these results suggest that current iEEG FMs do not yet transfer effectively to these continuous regression tasks, even with substantial adaptation.

The BrainTreebank results further show that this limitation is task-dependent. For this sEEG language classification task, the same iEEG FMs perform substantially better, with PopT reaching an AUROC of $0.779$ on the established word-vs.-non-word task and remaining competitive with CORTEG. CORTEG also generalizes to this setting and retains a clear benefit from EEG pretraining. Therefore, these results suggest that transfer performance depends on how well the pretrained representation matches the downstream task and on the adaptation strategy.

\paragraph{What are the advantages of CORTEG?}
CORTEG's main advantage is not only higher mean correlation, but also a practical deployment strategy based on adapting a pretrained cross-patient model. With LOO-FT, a new patient can be calibrated by only performing the Stage~2 fine-tuning of the existing population model, reducing adaptation time to roughly $10$--$30$ minutes on a single GPU, rather than retraining the pooled model for hours. The resulting model closely approaches pooled performance on both tasks and, under a $|\Delta r|{<}0.05$ tie convention, matches or exceeds per-subject training on all $25$ subjects in our benchmark. At the same time, pooled CORTEG achieves the highest mean correlation among the compared methods on both tasks, with modest gains on the public finger benchmark and larger, statistically significant gains on audio. This adaptation is also parameter-efficient, updating only $297$K parameters through LoRA.

\paragraph{Limitations} \textit{(i) MNI-space localization.} The Stanford dataset shows large across-subject variance (Fig.~\ref{fig:decoding}a), consistent with heterogeneous implant coverage. Identical MNI coordinates do not consistently correspond to identical function, and our shared-space adapter cannot fully exploit subject-specific anatomy. Subject-conditioned adapters or functional-connectivity priors are a natural next step. \textit{(ii) No cross-task transfer.} We evaluate scalp-EEG$\rightarrow$ECoG transfer separately for each task, but do not transfer a trained decoder across ECoG tasks (finger$\leftrightarrow$audio). BrainTreebank extends the evaluation to a new dataset, intracranial modality, and classification task, but represents generalization of the transfer framework rather than direct cross-task transfer. \textit{(iii) No scaling law.} Model scaling is not observed (Fig.~\ref{fig:decoding}b). This may reflect the limited amount of patient-specific ECoG data rather than an architectural ceiling, and joint scaling of pretraining and calibration data remains an open direction. \textit{(iv) Fusion remains open.} Mean-pooling provides a simple parameter-free default, while most alternative fusion mechanisms do not consistently improve it (App.~\ref{app:fusion}). Layer-wise gated fusion avoids selecting a fixed merge point and improves several Base/Large configurations, but its benefit is not consistent across scales (Table~\ref{tab:gate_main}; App.~\ref{app:gate}). Developing a fusion strategy that improves consistently across tasks and model scales remains future work.

\paragraph{Broader impact}
ECoG BCIs target patients with severe motor or communication impairment. Reducing per-patient calibration time lowers the deployment barrier but may introduce concentration risk, as a single population model affects many patients and may propagate biases present in the pretraining data.

%=============================================================================
\section{Conclusion}
\label{sec:conclusion}
%=============================================================================

CORTEG combines a pretrained EEG FM backbone, a coordinate-aware spatial adapter, dual-stream tokenization, and parameter-efficient adaptation to bridge scalp recordings and intracranial decoding. Across motor and auditory tasks, it achieves the highest mean performance among the compared methods while allowing a population model to be adapted to a held-out patient in minutes rather than retrained from scratch. The benefit of EEG pretraining is task- and backbone-dependent, but also generalizes to a public sEEG classification setting on BrainTreebank. Meanwhile, the contrasting performance of intracranial FMs across regression and classification highlights the importance of the downstream task, pretrained representation, and model-specific adaptation. Overall, our results support EEG foundation models as a useful source of transferable representations for ECoG and motivate further work on subject-specific anatomy, cross-task transfer, and improved adaptation strategies.

\begin{ack}
\text{*}L.Y. is supported by the Research Foundation – Flanders (FWO) grant 1S65622N.\\
\text{*}Q.S. is supported by the China Scholarship Council (no. 202206050022).\\
\text{*}B.V.D. is Belgian Fund for Scientific Research -
Flanders grant G0C1522N.\\
\text{*}E.C.M. is supported by the Research Foundation – Flanders (FWO) grant 1102925N.\\
\text{*}M.M.V.H. is supported by research grants received from Horizon Europe's Marie Sklodowska-Curie Action (grant agreement No. 101118964), Horizon 2020 research and innovation programme under grant agreement No. 857375, the special research fund of the KU Leuven (C24/18/098), the Belgian
Fund for Scientific Research -- Flanders (G0A4321N, G0C1522N, G031426N), and the Hercules Foundation (AKUL 043).\\
\text{*}The resources and services used in this work were provided by the VSC (Flemish Supercomputer Center), funded by the Research Foundation - Flanders (FWO) and the Flemish Government.\\
\text{*}We thank the neurosurgery and neurology team at Ghent University Hospital — in particular Dr. Alfred Meurs, Dr. Evelien Carrette, Dr. Dirk Van Roost, and Kato Van Rooy — for performing the electrode implantations and for their assistance in recording the intracranial audio-envelope data from the participating patients. We are especially grateful to the patients themselves for agreeing to take part in this research during their epilepsy-monitoring stay.
\end{ack}

\bibliographystyle{unsrtnat}
\bibliography{references}

%%%%%%%%%%%%%%%%%%%%%%%%%%%%%%%%%%%%%%%%%%%%%%%%%%%%%%%%%%%%
%=============================================================================
\newpage
\appendix
\section{Appendix}
%=============================================================================

%-----------------------------------------------------------------------------
\subsection{LOO-FT Strategy Sweep and Failed Alternatives}
\label{app:recipe}
%-----------------------------------------------------------------------------

\paragraph{Loss and metric} CORTEG minimizes per-subject z-scored MSE
\begin{equation}
    \mathcal{L} = \tfrac{1}{N}\sum_{i=1}^{N} \|\hat{y}^{(i)} - y^{(i)}\|_2^2,\qquad y^{(i)} = (y^{(i)}_\text{raw} - \mu_s)/\sigma_s,
    \label{eq:loss}
\end{equation}
and we report subject-level Pearson correlation
\begin{equation}
    \bar{r}_s = \tfrac{1}{d_\text{out}}\sum_{d=1}^{d_\text{out}} r_d^{(s)},\qquad \bar{r} = \tfrac{1}{|\mathcal{S}|}\sum_{s} \bar{r}_s,
    \label{eq:corr}
\end{equation}

We compared three Stage-2 protocols on the finger task ($n{=}9$); all share Stage~1 (pooled on $N{-}1$ subjects) and differ only in what is trainable in Stage~2. The results for different Stage-2 strategies are summarized in Table~\ref{tab:recipe_sweep}.

\begin{table}[ht]
\caption{LOO-FT Stage-2 protocols on the finger task ($f{=}1.0$, $n{=}9$, paired Wilcoxon vs.\ the pooled-training upper bound $r{=}0.554$). v3 is the recipe used in the main paper.}
\label{tab:recipe_sweep}
\centering
\small
\begin{tabular}{@{}llcc@{}}
\toprule
\textbf{Variant} & \textbf{Stage-2 trainable modules} & \textbf{Mean $r$} & \textbf{$p$ vs.\ pooled} \\
\midrule
v1          & full-backbone fine-tune (lr=$10^{-4}$)                                                & $\sim0.27$          & catastrophic overfit \\
v2          & head + LoRA + adapter, single base LR                                                & $0.5309$            & $0.301$ \\
\textbf{v3} & v2 + \textbf{adapter LR boost} (10$\times$ at $f{\geq}0.25$, 2$\times$ at $f{=}0.1$) & $\mathbf{0.5512}$   & $0.652$ \\
\bottomrule
\end{tabular}
\end{table}

The full-backbone variant (v1) collapses because the $\sim$25M backbone parameters far exceed the held-out subject's $\sim$$9$K Stage-2 training windows ($\sim$$40$K window-finger targets). Parameter-efficient adaptation (LoRA on the last 4 blocks) is therefore essential. Within the parameter-efficient regime, going from v2 to v3 (adapter LR boost) provides a further $0.02$ correlation gain by letting the spatial adapter quickly escape the Stage-1 LOO bias on the held-out subject's channel layout, while keeping LoRA at the base rate so the population prior is not overwritten.

\paragraph{Final strategy (v3)} The strategy used throughout the main paper combines: (i)~a pooled Stage~1 on $N{-}1$ subjects (AdamW, learning rate $3{\times}10^{-3}$, patience $90$, $100$ epochs, batch $64$, weight decay $5{\times}10^{-3}$; on audio, learning rate $10^{-3}$ and patience $50$). (ii)~a per-subject Stage~2 on the held-out subject's training split (AdamW, base LR $1{\times}10^{-3}$, patience $30$, $100$ epochs, batch $16$, weight decay $5{\times}10^{-3}$) with a fraction-aware adapter LR multiplier that raises the spatial adapter's effective learning rate to $10\times$ the base when $\geq{}25\%$ of the held-out subject's recording is available, and to $2\times$ at the most data-scarce setting ($f{=}0.1$). The asymmetry --- LoRA at the base rate, adapter at $10\times$ is because the spatial adapter depends on subject-specific channel layout, so it must adapt fastest in Stage~2, while LoRA's role is to refine an attention prior that already transfers across subjects and should not be overwritten. v3 closes the gap to the pooled upper bound ($r{=}0.554$) to within paired-Wilcoxon noise ($r{=}0.5512$, $p{=}0.652$), demonstrating that LOO-FT can recover pooled-level decoding without ever having seen the held-out subject during Stage~1.

          % A.1   app:recipe
%-----------------------------------------------------------------------------
\subsection{Dataset Preprocessing}
\label{app:data_prep}
%-----------------------------------------------------------------------------

Throughout the appendix, we refer to the two ECoG datasets by their recording sites: \emph{Stanford} is the public finger-trajectory corpus~\cite{miller2019library} and \emph{Ghent} is the private audio-envelope corpus described below.
%-----------------------------------------------------------------------------
\subsubsection{Finger trajectory regression}
The Stanford \textit{fingerflex} dataset \citep{miller2019library} contains recordings from nine epilepsy patients implanted with subdural ECoG grids (4 mm diameter, 1 cm pitch) across various cortical regions\footnote{Ethics statement: All patients participated in a purely voluntary manner, after providing informed written consent, under experimental protocols approved by the Institutional Review Board of the University of Washington (\#12193). All patient data were anonymized according to IRB protocol, in accordance with HIPAA mandate. These data originally appeared in the manuscript “Human Motor Cortical Activity Is Selectively Phase- Entrained on Underlying Rhythms” published in PLoS Computational Biology in 2012 \citep{miller2012human}}. Participants performed self-paced flexion-extension of individual fingers in response to visual cues. At the same time, ECoG signals (band-pass filtered at 0.3–200 Hz) and finger trajectories (recorded via a digital glove) were sampled at 1000 Hz and 25 Hz, respectively.

For signal preprocessing, we first applied a one-pass, 3rd-order Butterworth IIR band-stop filter to the continuous ECoG signals to eliminate power line interference at 60 Hz and its harmonics (120 Hz, 180 Hz). Then, we performed visual inspection to exclude abnormal channels, and finally referenced the signals to the common average (CAR). Data were partitioned chronologically in line with \citep{lin2025towards, sun2025spectro}: for recordings exceeding 600 seconds, the first 400 seconds served as the training set; otherwise, a 2:1 train-and-test split was applied. Within the training set, we kept the last 10\% of the data as the validation set.

The HGA stream was obtained by a fourth-order Butterworth bandpass filter at $70$--$200$\,Hz followed by Hilbert envelope extraction and average-pool decimation to $200$\,Hz; the LFS stream was obtained by a bandpass filter at $1$--$64$\,Hz with anti-aliasing decimation to $128$\,Hz. The two streams use the band ranges quoted in Fig.~\ref{fig:overview}; the audio task uses a slightly narrower HGA band (60--124\,Hz) because of its lower native sampling rate, as noted below.

Training, validation, and testing datasets were constructed using a sliding window approach. To strictly maintain causality, each input sample comprises a 1-second window of ECoG signals immediately preceding the movement. The corresponding label for each sample represents the subsequent trajectory positions of the five fingers. We applied a 40 ms sliding stride to match the 25 Hz sampling rate of the motion-tracking glove.

\subsubsection{Audio envelope regression}
ECoG data were collected from 16 patients undergoing presurgical evaluation for drug-resistant focal epilepsy at Ghent University Hospital after approval from its Commission Medical Ethics. All participants, or their legal guardians if they were under 18, signed the informed consent form before the experiments. Electrode placement was determined solely by clinical need, and neural activity was recorded using subdural grids and strips with a Micromed SD LTM 64 Express system. Participants passively listened to a 30-minute fairy tale presented in three blocks, with comprehension questions between blocks. 

The raw ECoG signals were sampled at $256$\,Hz. Bad channels were rejected automatically by a per-channel variance heuristic across the full recording (channels with std$\,{<}\,10^{-6}$, or whose variance exceeded $5\times$ or fell below $0.2\times$ the across-channel median). The retained channels were then re-referenced to the common average and z-scored within each of the three recording blocks, with statistics computed on that block's training portion only. A $50$\,ms stimulus-to-cortex delay was applied by shifting the envelope relative to the ECoG. The HGA stream was obtained by a fourth-order Butterworth bandpass at $60$--$124$\,Hz followed by Hilbert envelope extraction and resampling to $200$\,Hz; the LFS stream was decimated to $128$\,Hz. The HGA upper edge here is capped a few Hz below the Nyquist frequency of the $256$\,Hz native sampling rate, and is therefore narrower than the generic ``70--200\,Hz'' high-gamma band described in Fig.~\ref{fig:overview}, which corresponds to the Stanford recordings sampled at $1$\,kHz. The broadband speech envelope was extracted with a $28$-channel ERB-spaced gammatone filterbank between $50$\,Hz and $5$\,kHz~\cite{sondergaard2013auditory}; subband amplitudes were rectified, compressed with a $0.6$ power law, averaged, and resampled to $128$/$200$\,Hz to match the two ECoG streams. Each block was split chronologically into $80\%$/$10\%$/$10\%$ train/val/test, and the per-block splits were concatenated across blocks to form each subject's final arrays. For evaluation, the LFS/HGA streams and the broadband envelope are sliding-windowed at a $50$\,ms stride (CORTEG and the iEEG FMs train on windows at a $0.2$\,s step); each $1$\,s window predicts one envelope value taken at the right edge of the window, so the effective output rate is $20$\,Hz, matching the $0.05$\,s stride quoted in \S\ref{sec:setup}.

\paragraph{Fair-comparison protocol} For all neural baselines on the two regression tasks, we use a single shared protocol: identical train / val / test splits per task, identical per-subject z-score normalization (the iEEG FMs instead apply their own preprocessing to the raw input), identical evaluation windows (1-s lookback, 0.04 s / 0.05 s evaluation stride for finger/audio), and validation-driven early stopping. Exceptions to the 1-s lookback are Brant ($6$\,s; App.~\ref{app:fm}) and DeepFingerNet on Ghent ($2$\,s). Training strides, epochs, and patience are model-specific.

%-----------------------------------------------------------------------------
\subsubsection{BrainTreebank}
\label{app:btb}
%-----------------------------------------------------------------------------

Section~\ref{sec:results:btb} reports the BrainTreebank~\cite{wang2024braintreebank} classification results. Here we describe the corresponding implementation details.

\paragraph{Dataset description}
We use $10$ subjects with one session per subject (Table~\ref{tab:btb_cohort}). For seven subjects, we follow the held-out subject/session selection used by PopT and BrainBERT. \texttt{sub\_5}, \texttt{sub\_8}, and \texttt{sub\_9} are not included in that split and therefore contribute their only available session. CORTEG and the from-scratch baselines use PopT's \texttt{clean\_laplacian} electrode list intersected with the electrodes having MNI coordinates. The raw spectral baselines use the corresponding per-subject voxel table. We also note that \texttt{sub\_5}, \texttt{sub\_8}, and \texttt{sub\_9} were included in the BrainBERT and PopT pretraining dataset.

\begin{table}[ht]
\caption{BrainTreebank subject review. Three subjects (\texttt{sub\_3}, \texttt{sub\_7}, and \texttt{sub\_10}) watched the same film (\emph{Cars 2}) and therefore share highly similar event timings.}
\label{tab:btb_cohort}
\centering
\footnotesize
\setlength{\tabcolsep}{6pt}
\begin{tabular}{@{}llll@{}}
\toprule
\textbf{Subject} & \textbf{Trial} & \textbf{Stimulus film} & \textbf{Note} \\
\midrule
\texttt{sub\_1}  & trial001 & the-martian                & PopT/BrainBERT held-out split \\
\texttt{sub\_2}  & trial006 & aquaman                    & PopT/BrainBERT held-out split \\
\texttt{sub\_3}  & trial000 & cars-2                     & shared stimulus \\
\texttt{sub\_4}  & trial000 & shrek-the-third            & PopT/BrainBERT held-out split \\
\texttt{sub\_5}  & trial000 & fantastic-mr-fox           & in FM pretraining data \\
\texttt{sub\_6}  & trial004 & coraline                   & PopT/BrainBERT held-out split \\
\texttt{sub\_7}  & trial000 & cars-2                     & shared stimulus \\
\texttt{sub\_8}  & trial000 & sesame-street-episode-3990 & in FM pretraining data \\
\texttt{sub\_9}  & trial000 & ant-man                    & in FM pretraining data \\
\texttt{sub\_10} & trial000 & cars-2                     & shared stimulus \\
\bottomrule
\end{tabular}
\end{table}

\paragraph{Event construction}
We evaluate two binary language-classification tasks. Task~A distinguishes sentence-initial from mid-sentence words. Task~B is the established word-vs.\ non-word benchmark used by BrainBERT and PopT. Following their implementation, positive events are word onsets and negative events are non-overlapping $1$\,s intervals containing no word. The two classes are balanced by subsampling to the size of the smaller class using a fixed seed, with a maximum of $900$ events per class for computational efficiency.

For Task~B, we follow the same negative-sampling procedure used in BrainBERT and PopT, with one modification: negative events are restricted to the valid movie trigger period, and any events outside this range are removed before splitting the data. Because the original negatives are defined using $1$ s speech-free intervals, while some models use longer input windows, the resulting windows can still overlap or include nearby speech. Many negative events also occur during long silent periods. We therefore include Task~A as a complementary benchmark, where both classes are words.

\paragraph{Model-specific input windows}
Input windows follow the requirements of each model. CORTEG and the from-scratch baselines use $[t,t{+}1.5]$\,s windows. On Task~B, the upstream BrainBERT and PopT configurations use $[t{-}2.5,t{+}2.5]$\,s context. Brant uses its native $6$\,s patch, ending at the right edge of the post-onset window on Task~A ($[t{-}4.5,t{+}1.5]$) and of the BrainBERT/PopT window on Task~B ($[t{-}3.5,t{+}2.5]$). For Task~A, only post-onset activity is used because the pause before a sentence-initial word is already highly predictive of the label: pause duration alone reaches $0.85$--$0.93$ AUROC without neural data. This avoids turning the task into a pause-detection problem.

Different models use different input-window lengths, but all use the same train/validation/test splits. We exclude Brant's $L{=}15$ configuration because its approximately $90$ s input span would require an excessively large gap and leave too little usable data.

\paragraph{Data splits}
Randomly sampled BrainTreebank splits can leak information across temporally adjacent events~\cite{han2025diver}, and recent benchmarks therefore adopt temporally separated evaluation~\cite{zahorodnii2025neuroprobe,oganesian2026barista,han2025diver}. We therefore construct new chronological splits with explicit separation between training, validation, and test data. Moreover, BrainTreebank consists of continuous recordings, and nearby events can produce overlapping analysis windows. We use a $7$\,s gap between blocks to prevent this overlap. The gap is longer than the largest input window used in our comparison, and the same splits are used for all models. Each session is divided into $n_\text{folds}{+}1$ chronological blocks. For fold $i$, earlier blocks are used for fitting and validation, and the following block is used for testing:
\[
\texttt{fit} \;|\; 7\text{s gap} \;|\; \texttt{validation}
\;|\; 7\text{s gap} \;|\; \texttt{test}
\]
We use four folds for all models and both tasks, with $15\%$ of the pre-test data reserved for validation. Thus, every test block occurs strictly after the data used for training and model selection.

\paragraph{Model-specific settings}
We preserve each model's intended preprocessing and adaptation strategy where possible. CORTEG uses the $70$--$200$\,Hz high-gamma band together with MNI coordinates and a subject-specific low-rank adaptation. BrainBERT and PopT use Laplacian-referenced inputs, whereas Brant and the raw baselines use monopolar recordings.
  % A.2   app:data_prep
%-----------------------------------------------------------------------------
\subsection{Full Per-Subject Results}
\label{app:per_subject_results}
%-----------------------------------------------------------------------------

We report per-subject Pearson $r$ for every benchmarked method on both datasets. Subject identifiers follow the same anonymization as Fig.~\ref{fig:implant_finger}/\ref{fig:implant_audio} (mapping in the table captions). Table~\ref{tab:per_subject_finger_full} gives the full per-subject \emph{and} per-finger breakdown for the Stanford finger task. Table~\ref{tab:per_subject_audio_all} gives the per-subject breakdown for the audio-envelope task (single $r$ per subject since the target is a one-dimensional envelope). Table~\ref{tab:btb_persubj} gives the per-subject breakdown for the two BrainTreebank classification tasks.

{\footnotesize
\setlength{\tabcolsep}{3pt}
\renewcommand{\arraystretch}{1.1}

\begin{xltabular}{\linewidth}{@{}l@{\hspace{4pt}}l*{10}{Y}@{}}

\caption{Stanford finger regression task --- full per-subject \emph{and} per-finger Pearson $r$ for every benchmarked method, $n{=}9$, $f{=}1.0$. Each method block has five rows (Thumb, Index, Middle, Ring, Little) plus an \emph{Avg} row showing the per-subject mean across the five fingers; the rightmost column is the per-finger mean across the nine subjects. In the \emph{Avg} row, per-column \textcolor{blue}{\textbf{best}} (bold blue) and \textbf{second-best} (bold black) values are highlighted, matching the convention of Table~\ref{tab:main}. Rows marked $^{\dagger}$ are transcribed from~\cite{sun2025spectro} Table~V; rows marked $^{\ddagger}$ from~\cite{tao2025deepfingernet} Table~II. All other rows are computed from local outputs. Subject mapping: S1=bp, S2=cc, S3=ht, S4=jc, S5=jp, S6=mv, S7=wc, S8=wm, S9=zt (alphabetical, matching~\cite{sun2025spectro,tao2025deepfingernet}). Per-method grand means agree with Table~\ref{tab:main} within $\pm 0.004$ (small differences come from how the cited papers compute the grand average over per-finger vs.\ per-cell means).}
\label{tab:per_subject_finger_full}\\
\toprule
\textbf{Method} & \textbf{Finger} & \textbf{S1} & \textbf{S2} & \textbf{S3} & \textbf{S4} & \textbf{S5} & \textbf{S6} & \textbf{S7} & \textbf{S8} & \textbf{S9} & \textbf{Mean} \\
\midrule
\endfirsthead
\multicolumn{12}{l}{\textit{Table~\ref{tab:per_subject_finger_full} (continued)}}\\
\toprule
\textbf{Method} & \textbf{Finger} & \textbf{S1} & \textbf{S2} & \textbf{S3} & \textbf{S4} & \textbf{S5} & \textbf{S6} & \textbf{S7} & \textbf{S8} & \textbf{S9} & \textbf{Mean} \\
\midrule
\endhead
\bottomrule
\endfoot

\multirow[t]{6}{*}{Ridge\_LFS}
 & Thumb  & $0.306$ & $0.358$ & $0.058$ & $0.242$ & $0.127$ & $0.110$ & $0.162$ & $0.065$ & $0.244$ & $0.186$ \\
 & Index  & $0.250$ & $0.325$ & $0.179$ & $0.253$ & $0.172$ & $0.352$ & $0.231$ & $0.056$ & $0.190$ & $0.223$ \\
 & Middle & $0.209$ & $0.195$ & $0.034$ & $0.132$ & $0.041$ & $0.153$ & $0.171$ & $-0.014$ & $0.134$ & $0.117$ \\
 & Ring   & $0.338$ & $0.283$ & $0.262$ & $0.343$ & $0.061$ & $0.250$ & $0.166$ & $0.132$ & $0.232$ & $0.230$ \\
 & Little & $0.170$ & $0.281$ & $0.156$ & $0.325$ & $0.031$ & $0.219$ & $0.098$ & $-0.081$ & $0.161$ & $0.151$ \\
 & \textit{Avg} & $0.255$ & $0.288$ & $0.138$ & $0.259$ & $0.086$ & $0.217$ & $0.166$ & $0.032$ & $0.192$ & $0.181$ \\
\midrule

\multirow[t]{6}{*}{Ridge\_HGA}
 & Thumb  & $0.469$ & $0.594$ & $0.307$ & $0.506$ & $0.364$ & $0.225$ & $0.448$ & $0.250$ & $0.556$ & $0.413$ \\
 & Index  & $0.282$ & $0.441$ & $0.191$ & $0.436$ & $0.440$ & $0.588$ & $0.556$ & $0.042$ & $0.663$ & $0.404$ \\
 & Middle & $0.142$ & $0.483$ & $0.155$ & $0.253$ & $0.036$ & $0.190$ & $0.121$ & $0.228$ & $0.059$ & $0.185$ \\
 & Ring   & $0.403$ & $0.465$ & $0.253$ & $0.363$ & $0.531$ & $0.249$ & $0.234$ & $0.371$ & $0.400$ & $0.363$ \\
 & Little & $0.187$ & $0.609$ & $0.235$ & $0.252$ & $0.350$ & $0.580$ & $0.200$ & $0.083$ & $0.308$ & $0.312$ \\
 & \textit{Avg} & $0.296$ & $0.519$ & $0.229$ & $0.362$ & $0.344$ & $0.367$ & $0.312$ & $0.195$ & $0.397$ & $0.336$ \\
\midrule

\multirow[t]{6}{*}{PLS$^{\dagger}$}
 & Thumb  & $0.526$ & $0.640$ & $0.308$ & $0.556$ & $0.640$ & $0.481$ & $0.472$ & $0.256$ & $0.617$ & $0.500$ \\
 & Index  & $0.338$ & $0.429$ & $0.188$ & $0.419$ & $0.568$ & $0.725$ & $0.595$ & $0.133$ & $0.663$ & $0.451$ \\
 & Middle & $0.219$ & $0.485$ & $0.238$ & $0.309$ & $0.235$ & $0.265$ & $0.191$ & $0.228$ & $0.272$ & $0.271$ \\
 & Ring   & $0.434$ & $0.593$ & $0.321$ & $0.453$ & $0.607$ & $0.397$ & $0.302$ & $0.390$ & $0.499$ & $0.444$ \\
 & Little & $0.340$ & $0.618$ & $0.288$ & $0.316$ & $0.321$ & $0.640$ & $0.225$ & $-0.004$ & $0.340$ & $0.343$ \\
 & \textit{Avg} & $0.371$ & $0.553$ & $0.269$ & $0.411$ & $0.474$ & $0.502$ & $0.357$ & $0.201$ & $0.478$ & $0.402$ \\
\midrule

\multirow[t]{6}{*}{HOPLS$^{\dagger}$}
 & Thumb  & $0.501$ & $0.599$ & $0.328$ & $0.599$ & $0.367$ & $0.284$ & $0.494$ & $0.210$ & $0.634$ & $0.446$ \\
 & Index  & $0.337$ & $0.518$ & $0.248$ & $0.433$ & $0.328$ & $0.182$ & $0.610$ & $0.004$ & $0.695$ & $0.373$ \\
 & Middle & $0.212$ & $0.565$ & $0.259$ & $0.364$ & $0.149$ & $0.194$ & $0.115$ & $0.131$ & $0.288$ & $0.253$ \\
 & Ring   & $0.481$ & $0.581$ & $0.338$ & $0.422$ & $0.290$ & $0.522$ & $0.287$ & $0.469$ & $0.480$ & $0.430$ \\
 & Little & $0.305$ & $0.642$ & $0.374$ & $0.292$ & $0.198$ & $0.519$ & $0.171$ & $0.014$ & $0.394$ & $0.323$ \\
 & \textit{Avg} & $0.367$ & $0.581$ & $0.309$ & $0.422$ & $0.266$ & $0.340$ & $0.335$ & $0.166$ & $0.498$ & $0.365$ \\
\midrule

\multirow[t]{6}{*}{LSTM\_LFS}
 & Thumb  & $0.468$ & $0.597$ & $0.225$ & $0.627$ & $0.063$ & $0.441$ & $0.199$ & $0.063$ & $0.374$ & $0.340$ \\
 & Index  & $0.485$ & $0.497$ & $0.394$ & $0.452$ & $0.067$ & $0.315$ & $0.274$ & $-0.018$ & $0.209$ & $0.297$ \\
 & Middle & $0.399$ & $0.371$ & $0.157$ & $0.176$ & $0.090$ & $0.248$ & $0.235$ & $0.062$ & $0.203$ & $0.216$ \\
 & Ring   & $0.445$ & $0.526$ & $0.465$ & $0.496$ & $0.039$ & $0.188$ & $0.078$ & $0.069$ & $0.233$ & $0.282$ \\
 & Little & $0.350$ & $0.524$ & $0.315$ & $0.390$ & $-0.068$ & $0.141$ & $0.137$ & $0.064$ & $0.344$ & $0.244$ \\
 & \textit{Avg} & $0.429$ & $0.503$ & $0.311$ & $0.428$ & $0.038$ & $0.267$ & $0.185$ & $0.048$ & $0.273$ & $0.276$ \\
\midrule

\multirow[t]{6}{*}{LSTM\_HGA$^{\dagger}$}
 & Thumb  & $0.600$ & $0.745$ & $0.429$ & $0.668$ & $0.678$ & $0.531$ & $0.487$ & $0.217$ & $0.498$ & $0.539$ \\
 & Index  & $0.468$ & $0.639$ & $0.388$ & $0.568$ & $0.628$ & $0.864$ & $0.561$ & $0.126$ & $0.696$ & $0.549$ \\
 & Middle & $0.533$ & $0.608$ & $0.239$ & $0.429$ & $0.145$ & $0.562$ & $0.268$ & $0.420$ & $0.474$ & $0.409$ \\
 & Ring   & $0.572$ & $0.647$ & $0.425$ & $0.429$ & $0.673$ & $0.531$ & $0.272$ & $0.389$ & $0.545$ & $0.498$ \\
 & Little & $0.378$ & $0.703$ & $0.370$ & $0.383$ & $0.519$ & $0.776$ & $0.268$ & $0.097$ & $0.430$ & $0.436$ \\
 & \textit{Avg} & $0.510$ & $0.668$ & $0.370$ & $0.495$ & $0.529$ & $0.653$ & $0.371$ & $\mathbf{0.250}$ & $0.529$ & $0.486$ \\
\midrule

\multirow[t]{6}{*}{CNN-LSTM$^{\dagger}$}
 & Thumb  & $0.518$ & $0.702$ & $0.233$ & $0.553$ & $0.531$ & $0.508$ & $0.459$ & $0.319$ & $0.456$ & $0.475$ \\
 & Index  & $0.472$ & $0.577$ & $0.227$ & $0.435$ & $0.655$ & $0.707$ & $0.593$ & $0.111$ & $0.716$ & $0.499$ \\
 & Middle & $0.273$ & $0.520$ & $0.140$ & $0.407$ & $0.213$ & $0.216$ & $0.234$ & $0.226$ & $0.403$ & $0.292$ \\
 & Ring   & $0.496$ & $0.587$ & $0.331$ & $0.407$ & $0.559$ & $0.427$ & $0.275$ & $0.314$ & $0.549$ & $0.438$ \\
 & Little & $0.280$ & $0.571$ & $0.294$ & $0.361$ & $0.452$ & $0.528$ & $0.285$ & $0.058$ & $0.417$ & $0.361$ \\
 & \textit{Avg} & $0.408$ & $0.591$ & $0.245$ & $0.433$ & $0.482$ & $0.477$ & $0.369$ & $0.206$ & $0.508$ & $0.413$ \\
\midrule

\multirow[t]{6}{*}{DeepFingerNet$^{\ddagger}$}
 & Thumb  & $0.66$ & $0.79$ & $0.41$ & $0.60$ & $0.63$ & $0.57$ & $0.52$ & $0.48$ & $0.63$ & $0.59$ \\
 & Index  & $0.68$ & $0.70$ & $0.39$ & $0.47$ & $0.63$ & $0.87$ & $0.63$ & $0.29$ & $0.77$ & $0.60$ \\
 & Middle & $0.46$ & $0.62$ & $0.30$ & $0.39$ & $0.44$ & $0.56$ & $0.52$ & $0.40$ & $0.47$ & $0.46$ \\
 & Ring   & $0.60$ & $0.77$ & $0.46$ & $0.48$ & $0.65$ & $0.55$ & $0.26$ & $0.23$ & $0.58$ & $0.51$ \\
 & Little & $0.59$ & $0.77$ & $0.39$ & $0.53$ & $0.55$ & $0.75$ & $0.33$ & $0.30$ & $0.58$ & $0.53$ \\
 & \textit{Avg} & $0.60$ & \textcolor{blue}{$\mathbf{0.73}$} & $0.39$ & $0.49$ & $0.58$ & $\mathbf{0.66}$ & $0.45$ & \textcolor{blue}{$\mathbf{0.34}$} & $0.61$ & $0.54$ \\
\midrule

\multirow[t]{6}{*}{HiLoFuseNet$^{\dagger}$}
 & Thumb  & $0.687$ & $0.687$ & $0.453$ & $0.790$ & $0.758$ & $0.667$ & $0.464$ & $0.377$ & $0.592$ & $0.608$ \\
 & Index  & $0.567$ & $0.683$ & $0.486$ & $0.684$ & $0.736$ & $0.855$ & $0.609$ & $0.135$ & $0.701$ & $0.606$ \\
 & Middle & $0.513$ & $0.615$ & $0.288$ & $0.497$ & $0.365$ & $0.421$ & $0.373$ & $0.347$ & $0.473$ & $0.432$ \\
 & Ring   & $0.641$ & $0.685$ & $0.583$ & $0.606$ & $0.675$ & $0.439$ & $0.344$ & $0.315$ & $0.638$ & $0.547$ \\
 & Little & $0.393$ & $0.695$ & $0.436$ & $0.618$ & $0.431$ & $0.717$ & $0.221$ & $0.060$ & $0.536$ & $0.456$ \\
 & \textit{Avg} & $0.560$ & $0.673$ & $\mathbf{0.449}$ & \textcolor{blue}{$\mathbf{0.639}$} & $0.593$ & $0.620$ & $0.402$ & $0.247$ & $0.588$ & $0.530$ \\
\midrule

\multirow[t]{6}{*}{CORTEG (per-subject)}
 & Thumb  & $0.652$ & $0.778$ & $0.510$ & $0.627$ & $0.696$ & $0.323$ & $0.605$ & $0.146$ & $0.559$ & $0.544$ \\
 & Index  & $0.650$ & $0.695$ & $0.435$ & $0.580$ & $0.594$ & $0.777$ & $0.723$ & $0.099$ & $0.725$ & $0.587$ \\
 & Middle & $0.664$ & $0.610$ & $0.369$ & $0.406$ & $0.288$ & $0.528$ & $0.580$ & $0.457$ & $0.628$ & $0.503$ \\
 & Ring   & $0.684$ & $0.710$ & $0.496$ & $0.576$ & $0.554$ & $0.637$ & $0.474$ & $0.405$ & $0.666$ & $0.578$ \\
 & Little & $0.515$ & $0.749$ & $0.470$ & $0.534$ & $0.433$ & $0.812$ & $0.428$ & $-0.014$ & $0.438$ & $0.485$ \\
 & \textit{Avg} & \textcolor{blue}{$\mathbf{0.633}$} & $0.708$ & \textcolor{blue}{$\mathbf{0.456}$} & $0.545$ & $0.513$ & $0.615$ & \textcolor{blue}{$\mathbf{0.562}$} & $0.219$ & $0.603$ & $0.539$ \\
\midrule

\multirow[t]{6}{*}{CORTEG (LOO-FT)}
 & Thumb  & $0.688$ & $0.782$ & $0.511$ & $0.642$ & $0.769$ & $0.525$ & $0.616$ & $0.185$ & $0.622$ & $0.593$ \\
 & Index  & $0.589$ & $0.689$ & $0.437$ & $0.603$ & $0.721$ & $0.824$ & $0.748$ & $0.078$ & $0.746$ & $0.604$ \\
 & Middle & $0.579$ & $0.607$ & $0.296$ & $0.433$ & $0.373$ & $0.557$ & $0.541$ & $0.388$ & $0.638$ & $0.490$ \\
 & Ring   & $0.668$ & $0.718$ & $0.440$ & $0.586$ & $0.684$ & $0.586$ & $0.488$ & $0.382$ & $0.653$ & $0.578$ \\
 & Little & $0.510$ & $0.687$ & $0.412$ & $0.520$ & $0.511$ & $0.809$ & $0.399$ & $0.073$ & $0.490$ & $0.490$ \\
 & \textit{Avg} & $\mathbf{0.607}$ & $0.697$ & $0.419$ & $\mathbf{0.557}$ & $\mathbf{0.612}$ & $\mathbf{0.660}$ & $\mathbf{0.559}$ & $0.221$ & $\mathbf{0.630}$ & $\mathbf{0.551}$ \\
\midrule

\multirow[t]{6}{*}{\textbf{CORTEG (Pooled)}}
 & Thumb  & $0.630$ & $0.793$ & $0.459$ & $0.680$ & $0.723$ & $0.646$ & $0.555$ & $0.128$ & $0.661$ & $0.586$ \\
 & Index  & $0.647$ & $0.709$ & $0.435$ & $0.594$ & $0.702$ & $0.787$ & $0.710$ & $0.100$ & $0.758$ & $0.605$ \\
 & Middle & $0.508$ & $0.619$ & $0.287$ & $0.344$ & $0.478$ & $0.728$ & $0.569$ & $0.452$ & $0.636$ & $0.513$ \\
 & Ring   & $0.614$ & $0.753$ & $0.422$ & $0.504$ & $0.630$ & $0.639$ & $0.418$ & $0.350$ & $0.627$ & $0.551$ \\
 & Little & $0.493$ & $0.704$ & $0.492$ & $0.501$ & $0.634$ & $0.764$ & $0.407$ & $0.124$ & $0.494$ & $0.513$ \\
 & \textit{Avg} & $0.578$ & $\mathbf{0.716}$ & $0.419$ & $0.525$ & \textcolor{blue}{$\mathbf{0.634}$} & \textcolor{blue}{$\mathbf{0.713}$} & $0.531$ & $0.231$ & \textcolor{blue}{$\mathbf{0.635}$} & \textcolor{blue}{$\mathbf{0.554}$} \\

\end{xltabular}
}

\begin{table}[ht]
\caption{Audio-envelope task --- per-subject Pearson $r$, $n{=}16$. Per-column \textcolor{blue}{\textbf{best}} (bold blue) and \textbf{second-best} (bold black) values are highlighted, matching the convention of Table~\ref{tab:main}. The grand-mean column matches Table~\ref{tab:main} exactly. Subject mapping: S1=2018\_001, S2=2019\_001, S3=2019\_003, S4=2019\_004, S5=2019\_007, S6=2020\_001, S7=2020\_002, S8=2020\_004, S9=2020\_005, S10=2020\_006, S11=2021\_001, S12=2021\_002, S13=2021\_005-1, S14=2021\_006, S15=2021\_007, S16=2021\_008-1. Ridge baselines are evaluated continuously (all time points) rather than on the same windowed test split as the deep methods, so their per-subject decomposition is not directly comparable and is reported in Table~\ref{tab:main} (mean only) rather than here.}
\label{tab:per_subject_audio_all}
\centering
\scriptsize
\setlength{\tabcolsep}{2pt}
\resizebox{\textwidth}{!}{%
\begin{tabular}{@{}lcccccccccccccccc|c@{}}
\toprule
\textbf{Method} & \textbf{S1} & \textbf{S2} & \textbf{S3} & \textbf{S4} & \textbf{S5} & \textbf{S6} & \textbf{S7} & \textbf{S8} & \textbf{S9} & \textbf{S10} & \textbf{S11} & \textbf{S12} & \textbf{S13} & \textbf{S14} & \textbf{S15} & \textbf{S16} & \textbf{Mean} \\
\midrule
LSTM\_LFS       & $0.152$ & $0.079$ & $0.261$ & $0.045$ & $0.336$ & $0.281$ & $0.065$ & $0.444$ & $0.061$ & $0.050$ & $0.290$ & $0.408$ & $0.085$ & $0.074$ & $0.025$ & $0.506$ & $0.198$ \\
LSTM\_HGA       & $-0.003$ & $0.012$ & $0.058$ & $0.014$ & $0.209$ & $0.143$ & $0.051$ & \textcolor{blue}{$\mathbf{0.576}$} & $0.059$ & $0.074$ & $0.083$ & $0.383$ & $0.050$ & $0.046$ & $0.050$ & $0.386$ & $0.137$ \\
CNN-LSTM        & $0.143$ & $0.040$ & $0.380$ & $0.103$ & $0.404$ & $0.349$ & $0.104$ & $0.550$ & $\mathbf{0.112}$ & $0.115$ & $0.346$ & $0.542$ & $0.097$ & $0.144$ & $0.115$ & $0.639$ & $0.261$ \\
DeepFingerNet   & $0.028$ & $0.004$ & $-0.002$ & $0.017$ & $0.051$ & $0.053$ & $0.043$ & $0.369$ & $0.050$ & $-0.009$ & $-0.001$ & $0.317$ & $-0.018$ & $0.044$ & $-0.005$ & $0.418$ & $0.085$ \\
HiLoFuseNet     & $0.109$ & $0.084$ & $0.307$ & $0.183$ & $\mathbf{0.493}$ & $0.334$ & $0.128$ & $0.562$ & $0.106$ & $0.072$ & $0.306$ & $0.560$ & $0.062$ & $0.140$ & $0.065$ & $0.640$ & $0.259$ \\
HOPLS           & $0.021$ & $0.023$ & $0.234$ & $0.090$ & $0.290$ & $0.325$ & \textcolor{blue}{$\mathbf{0.351}$} & $0.394$ & $0.098$ & $0.135$ & $0.176$ & $0.363$ & $0.158$ & $0.115$ & $0.026$ & $0.450$ & $0.203$ \\
PLS             & $0.035$ & $0.037$ & $0.263$ & $0.075$ & $0.290$ & $0.321$ & $\mathbf{0.339}$ & $0.403$ & $0.110$ & $0.087$ & $0.168$ & $0.402$ & $0.149$ & $0.079$ & $0.015$ & $0.467$ & $0.203$ \\
\midrule
CORTEG (per-subject)  & $-0.004$ & $0.052$ & $0.378$ & $0.024$ & $0.469$ & $0.237$ & $0.248$ & $0.568$ & $0.010$ & $0.210$ & $0.337$ & $0.557$ & $-0.053$ & \textcolor{blue}{$\mathbf{0.199}$} & $0.104$ & $\mathbf{0.658}$ & $0.250$ \\
CORTEG (LOO-FT)       & $\mathbf{0.164}$ & $\mathbf{0.116}$ & \textcolor{blue}{$\mathbf{0.423}$} & $\mathbf{0.194}$ & \textcolor{blue}{$\mathbf{0.507}$} & $\mathbf{0.431}$ & $0.327$ & $0.563$ & $0.064$ & \textcolor{blue}{$\mathbf{0.273}$} & \textcolor{blue}{$\mathbf{0.393}$} & \textcolor{blue}{$\mathbf{0.587}$} & \textcolor{blue}{$\mathbf{0.194}$} & $0.190$ & \textcolor{blue}{$\mathbf{0.199}$} & \textcolor{blue}{$\mathbf{0.663}$} & $\mathbf{0.331}$ \\
\textbf{CORTEG (Pooled)} & \textcolor{blue}{$\mathbf{0.230}$} & \textcolor{blue}{$\mathbf{0.125}$} & \textcolor{blue}{$\mathbf{0.423}$} & \textcolor{blue}{$\mathbf{0.226}$} & $\mathbf{0.493}$ & \textcolor{blue}{$\mathbf{0.433}$} & $0.337$ & $\mathbf{0.575}$ & \textcolor{blue}{$\mathbf{0.167}$} & $\mathbf{0.248}$ & $\mathbf{0.369}$ & $\mathbf{0.583}$ & $\mathbf{0.176}$ & $\mathbf{0.198}$ & $\mathbf{0.187}$ & $0.654$ & \textcolor{blue}{$\mathbf{0.339}$} \\
\bottomrule
\end{tabular}%
}
\end{table}

The variance across subjects can reflect clinical-grid heterogeneity (electrode count, hemisphere, sensorimotor vs.\ peri-Sylvian coverage; see App.~\ref{app:implantations}). On the finger task, every subject is decoded above chance by all CORTEG variants, and the per-subject ranking is consistent with the linear baselines' difficulty ordering, e.g.\ S8 (wm) is the hardest case for eleven of the twelve methods (the exception is LSTM\_LFS, where S5 is marginally lower, $0.038$ vs.\ $0.048$). On the audio task, per-subject CORTEG occasionally yields negative correlations (S1 and S13), whereas pooled CORTEG achieves (r>0.1) for every subject and LOO-FT for all but one (S9, r=0.064). This suggests that the population prior helps stabilize decoding for subjects whose limited individual data are insufficient to learn a reliable mapping.

\paragraph{DeepFingerNet on the audio task} DeepFingerNet's audio result ($r{=}0.085$, the lowest of any deep baseline) is surprising, because the same architecture is the headline state-of-the-art on the Stanford finger task (Table~\ref{tab:per_subject_finger_full}, $r{=}0.54$). Three reasons may explain it. First, the architecture is specifically designed and validated for finger-trajectory decoding only~\cite{tao2025deepfingernet}: the original paper evaluates only on BCIIV4 and Stanford, and the nested 3-UNet hierarchy plus 40-frequency Morlet-wavelet input are inductive biases tuned to the smooth 5-DoF kinematic targets typical of those datasets, not to envelope tracking. Second, the published numbers in~\cite{tao2025deepfingernet} are obtained on a 65/35 train/validation split where the validation set serves both for early stopping and as the reported metric, whereas our protocol holds out a separate windowed test split (matched to every other method in this paper) for a strictly fair cross-method comparison. Third, the model's 1.16M parameters, which is the largest among our baseline models, a 3.5$\times$ increase over HiLoFuseNet's 334K, combined with the dense wavelet representation, give the network ample capacity to memorize the small per-subject audio splits. Meanwhile, smaller dual-stream baselines (HiLoFuseNet, CNN-LSTM) and CORTEG (which freezes the bulk of the backbone via LoRA) all generalize better in this task.

\begin{table}[ht]
\caption{BrainTreebank per-subject AUROC.
Per-subject results for the methods in Table~\ref{tab:btb_main}, using each method's best evaluated configuration. \textbf{S}$i$ denotes \texttt{sub\_}$i$ in Table~\ref{tab:btb_cohort}. Within each task, the best result is shown in \textcolor{blue}{\textbf{bold blue}} and the second-best in \textbf{bold}. The \textbf{Mean} column reports the cohort mean $\pm$ cross-subject SD and matches Table~\ref{tab:btb_main}. CORTEG, its random-init control, and HiLoFuseNet are averaged over training seeds $1$, $2$, and $42$ on one event set drawn with seed $42$. PopT uses seed $42$, while BrainBERT and Brant use deterministic frozen probes. $^{\dagger}$~denotes the single-best-electrode arm, whose electrode is selected on the same folds that score it, giving a null of ${\approx}0.53$ rather than $0.50$.}
\label{tab:btb_persubj}
\centering
\scriptsize
\setlength{\tabcolsep}{2.5pt}
\resizebox{\textwidth}{!}{%
\begin{tabular}{@{}lcccccccccc|c@{}}
\toprule
\textbf{Method} & \textbf{S1} & \textbf{S2} & \textbf{S3} & \textbf{S4} & \textbf{S5} & \textbf{S6} & \textbf{S7} & \textbf{S8} & \textbf{S9} & \textbf{S10} & \textbf{Mean} \\
\midrule
\multicolumn{12}{@{}l}{\textcolor{gray}{\textit{Task~A: sentence-initial vs.\ mid-sentence word}}} \\[1pt]
BrainBERT$^{\dagger}$ & $0.577$ & $0.578$ & $0.616$ & $0.641$ & $\mathbf{0.680}$ & \textcolor{blue}{$\mathbf{0.634}$} & \textcolor{blue}{$\mathbf{0.546}$} & $\mathbf{0.614}$ & $\mathbf{0.532}$ & $0.598$ & $\mathbf{0.602{\pm}0.045}$ \\
PopT & $0.626$ & \textcolor{blue}{$\mathbf{0.644}$} & $0.613$ & \textcolor{blue}{$\mathbf{0.683}$} & \textcolor{blue}{$\mathbf{0.729}$} & $0.513$ & $0.502$ & $0.513$ & $0.510$ & \textcolor{blue}{$\mathbf{0.671}$} & $0.600{\pm}0.084$ \\
Brant$^{\dagger}$ & $\mathbf{0.651}$ & $0.545$ & $0.637$ & $0.605$ & $0.622$ & $0.563$ & $0.537$ & $0.596$ & \textcolor{blue}{$\mathbf{0.542}$} & $0.566$ & $0.587{\pm}0.041$ \\
HiLoFuseNet & $0.541$ & $0.521$ & $0.540$ & $0.558$ & $0.561$ & $0.530$ & $0.519$ & $0.555$ & $0.493$ & $0.514$ & $0.533{\pm}0.022$ \\
CORTEG, rand init & $0.556$ & $0.538$ & $\mathbf{0.682}$ & $0.510$ & $0.540$ & $0.517$ & $0.507$ & $0.503$ & $0.514$ & $0.509$ & $0.538{\pm}0.054$ \\
\textbf{CORTEG (ours)} & \textcolor{blue}{$\mathbf{0.745}$} & $\mathbf{0.627}$ & \textcolor{blue}{$\mathbf{0.756}$} & $\mathbf{0.671}$ & $0.636$ & $\mathbf{0.579}$ & $\mathbf{0.545}$ & \textcolor{blue}{$\mathbf{0.672}$} & $0.514$ & $\mathbf{0.631}$ & \textcolor{blue}{$\mathbf{0.638{\pm}0.078}$} \\
\midrule
\multicolumn{12}{@{}l}{\textcolor{gray}{\textit{Task~B: word vs.\ non-word}}} \\[1pt]
BrainBERT$^{\dagger}$ & $0.733$ & $0.719$ & $0.737$ & $0.678$ & $0.796$ & $\mathbf{0.873}$ & $0.619$ & $0.654$ & $0.552$ & $0.619$ & $0.698{\pm}0.094$ \\
PopT & $\mathbf{0.829}$ & \textcolor{blue}{$\mathbf{0.844}$} & $\mathbf{0.869}$ & $\mathbf{0.776}$ & \textcolor{blue}{$\mathbf{0.865}$} & \textcolor{blue}{$\mathbf{0.901}$} & $\mathbf{0.671}$ & $\mathbf{0.737}$ & \textcolor{blue}{$\mathbf{0.604}$} & \textcolor{blue}{$\mathbf{0.693}$} & \textcolor{blue}{$\mathbf{0.779{\pm}0.099}$} \\
Brant$^{\dagger}$ & $0.657$ & $0.556$ & $0.599$ & $0.559$ & $0.620$ & $0.544$ & $0.561$ & $0.574$ & $0.550$ & $0.555$ & $0.577{\pm}0.037$ \\
HiLoFuseNet & $0.756$ & $0.718$ & $0.863$ & $0.746$ & $\mathbf{0.821}$ & $0.721$ & \textcolor{blue}{$\mathbf{0.712}$} & $0.693$ & $0.549$ & $0.629$ & $0.721{\pm}0.089$ \\
CORTEG, rand init & $0.764$ & $0.590$ & $0.779$ & $0.586$ & $0.523$ & $0.562$ & $0.510$ & $0.560$ & $0.515$ & $0.489$ & $0.588{\pm}0.102$ \\
\textbf{CORTEG (ours)} & \textcolor{blue}{$\mathbf{0.912}$} & $\mathbf{0.802}$ & \textcolor{blue}{$\mathbf{0.891}$} & \textcolor{blue}{$\mathbf{0.834}$} & $0.625$ & $0.861$ & $0.550$ & \textcolor{blue}{$\mathbf{0.787}$} & $\mathbf{0.588}$ & $\mathbf{0.643}$ & $\mathbf{0.749{\pm}0.134}$ \\
\bottomrule
\end{tabular}}
\end{table}
    % A.3   app:per_subject_results
%-----------------------------------------------------------------------------
\subsection{Experimental Protocols for Figure~\ref{fig:decoding}}
\label{app:decoding_protocols}
%-----------------------------------------------------------------------------

This subsection describes the experimental protocol behind each panel of Figure~\ref{fig:decoding}. \emph{Reproducibility note:} the CORTEG-S score in Fig.~\ref{fig:decoding} (b) is higher than the $0.554$ in Table~\ref{tab:main} because it comes from the heavier HPC sweep with per-scale hyperparameter tuning. For consistency and reproducibility, Table~\ref{tab:main} reports only results obtained under our simple local benchmark (a single RTX 5090, $100$ epochs, batch size $64$ for pooled CORTEG, and the default merge point $K{=}4$). Per-subject CORTEG in Table~\ref{tab:main} uses a learning rate of $10^{-3}$, patience $30$ (audio $50$), and batch size $16$ on finger ($64$ on audio). Besides LoRA $\mathbf{A},\mathbf{B}$ ($131{,}072$), the spatial adapter ($142{,}095$), and the regression head ($2{,}565$ on finger), the $297$K trainable parameters of CORTEG-S include the high-gamma patch embedding ($13{,}312$) and the LayerNorms of the $K{=}4$ adapted blocks ($8{,}192$).

\paragraph{Panel (a): Full-data method comparison} All methods are trained at recording fraction $f{=}1.0$ (i.e.\ the full per-subject training split) and evaluated on the held-out test split. The reported quantity is the subject-level Pearson $r$ averaged across subjects ($n{=}9$ for finger; $n{=}16$ for audio). For the finger task, the per-subject $r$ is itself the mean over the five fingers. Bar heights are the cross-subject mean, and error bars are $\pm 1$ standard deviation across subjects. Per-method numbers are exactly those in Table~\ref{tab:main}; cited values from~\cite{sun2025spectro,tao2025deepfingernet} are used where available (marked $^{\dagger}$/$^{\ddagger}$ in App.~\ref{app:per_subject_results}). Stars above each non-pooled bar denote a two-sided paired Wilcoxon signed-rank test of per-subject $r$ against the CORTEG-pooled column, with Bonferroni correction across the four comparisons within each task (significance thresholds $p<0.05/0.01/0.001$ for ${*}/{**}/{***}$ and \emph{n.s.} otherwise. The three CORTEG bars (Pooled, LOO-FT, Per-subject) share the same architecture and trainable-parameter count and differ only in training regime; the LOO-FT bar uses the v3 recipe (App.~\ref{app:recipe}). Full per-subject and per-finger breakdowns of every method in this panel are in Tables~\ref{tab:per_subject_finger_full} and \ref{tab:per_subject_audio_all}.

\paragraph{Panel (b): Backbone scaling} We compare the three CORTEG architecture variants (Small, Base, Large; full hyperparameters in Table~\ref{tab:arch}) at pooled training under a matched reporting protocol: each violin reports per-subject Pearson $r$ extracted from the val-best (final saved checkpoint) per-subject table. We report the best-performing model based on a hyperparameter sweep on one factor: Merge point $K{=}\,$\texttt{hi\_inject\_last\_n}. We use the per-scale best from the merge sweep in Table~\ref{tab:merge}: Small $K{=}1$, Base $K{=}2$, Large $K{=}6$. This scaling experiment uses an HPC recipe (200 epochs, batch 128 with gradient accumulation 4 for an effective batch of 512, patience 190; learning rate $3{\times}10^{-3}$ for Small and Base, $1{\times}10^{-3}$ for Large because $3{\times}10^{-3}$ causes training collapse on the larger backbone) needed for stable convergence at scale. LoRA depth (last 4 of $L$) and rank are not re-tuned per scale: a deeper-LoRA sensitivity sweep on the Small backbone (App. Table~\ref{tab:lora_sens}) showed no improvement from extending LoRA to 6 or 8 layers. Each half-violin shows the kernel-density estimate of per-subject Pearson $r$ for the chosen configuration of that backbone, with the embedded box reporting the median and interquartile range; individual subject points are overlaid as dots. At the per-scale best operating point, CORTEG-Small and CORTEG-Base both achieve a paired-Wilcoxon-significant advantage over HiLoFuseNet on the finger task ($p{=}0.039$ and $p{=}0.027$ respectively, two-sided, $n{=}9$); the gap over DeepFingerNet ($+0.04$ absolute) trends in the same direction but does not reach $\alpha{=}0.05$ on $n{=}9$ subjects, which we attribute primarily to a power limit. CORTEG-Large is borderline ($p{=}0.074$).

\begin{table}[ht]
\caption{HPC-protocol finger-task statistical comparison against the strongest prior decoders. Per-subject correlations parsed from the HPC sweep logs at the per-scale best $K$ (Small $K{=}1$, Base $K{=}2$, Large $K{=}6$); two-sided paired Wilcoxon vs.\ HiLoFuseNet (App.~Table~\ref{tab:per_subject_finger_full}, transcribed from~\cite{sun2025spectro}) and DeepFingerNet (transcribed from~\cite{tao2025deepfingernet}, 2-decimal rounding). $n{=}9$ subjects. Significant entries ($p{<}0.05$) in \textbf{bold}.}
\label{tab:hpc_stats}
\centering
\small
\begin{tabular}{@{}lccc@{}}
\toprule
\textbf{HPC variant} & \textbf{Mean $r$ $\pm$ SD} & \textbf{vs HiLoFuseNet $p$} & \textbf{vs DeepFingerNet $p$} \\
\midrule
CORTEG-Small HPC & $0.577 \pm 0.157$ & $\mathbf{0.039}$ & $0.13$ \\
CORTEG-Base HPC  & $0.583 \pm 0.154$ & $\mathbf{0.027}$ & $0.13$ \\
CORTEG-Large HPC & $0.577 \pm 0.155$ & $0.074$          & $0.16$ \\
\bottomrule
\end{tabular}
\end{table}

\paragraph{Panel (c): Compute--accuracy trade-off} For each backbone, we report a single point: the $x$-coordinate is wall-clock training time for one pooled run measured end-to-end on a single RTX 5090, including warmup and early stopping, under the $100$-epoch local recipe of \S\ref{sec:setup} rather than the extended HPC schedule used for panel~(b); the $y$-coordinate is inference time per sample, and the matched grand-mean $r$ values from panel~(b) appear as text annotations beside each marker. The marker area is proportional to the total parameter count. The right-hand axis converts the inference-time scale into a prediction rate (Hz) using the per-window forward-pass cost averaged over $1000$ inferences after a $1000$-window warm-up (RTX 5090, fp32, batch 1).

\paragraph{Panels (d, e): Low-data scaling on finger and audio} Each subject's training-window pool is sub-sampled to a fraction $f \in \{0\%,\,10\%,\,25\%,\,50\%,\,100\%\}$ of the available recording. Held-out validation and test splits are kept at full size so that the metric is comparable across $f$. The five methods plotted (ridge regression on the high-gamma stream, i.e.\ Ridge\_HGA; HiLoFuseNet, CORTEG per-subject from a random-init backbone, CORTEG per-subject from the pretrained backbone, CORTEG LOO-FT) are each retrained from scratch at every fraction. The ridge series is re-run under this windowed low-data protocol rather than the full-data protocol of Table~\ref{tab:main}, so its $f{=}1.0$ point ($0.387$ finger, $0.136$ audio) differs from the Ridge\_HGA row there ($0.336$\,/\,$0.175$). All series are trained at every fraction, with all other hyperparameters held fixed at the $f{=}1.0$ values. The $f{=}0\%$ point shows only CORTEG LOO-FT (Stage~1 alone with no Stage~2 fine-tuning, i.e.\ a strict zero-shot bar) since the other methods cannot be defined at $f{=}0$. LOO-FT Stage~2 uses the fraction-aware adapter learning-rate boost from the v3 recipe ($10\times$ at $f{\geq}0.25$, $2\times$ at $f{=}0.1$; App.~\ref{app:recipe}). Bar height is the cross-subject mean, and error bars are $\pm 1$ standard deviation. Per-held-out-subject zero-shot ($f{=}0\%$) Pearson $r$ values, read from each LOO Stage-1 model's evaluation on its own held-out subject, are listed in Table~\ref{tab:zeroshot}.

\begin{table}[ht]
\caption{Per-held-out-subject zero-shot Pearson $r$ at $f{=}0\%$ (LOO Stage-1 only, no patient-specific tuning). Values are the held-out subject's $r$ recorded by the LOO Stage-1 evaluation pass; subject mapping matches App.~Table~\ref{tab:per_subject_finger_full} (finger) and App.~Table~\ref{tab:per_subject_audio_all} (audio).}
\label{tab:zeroshot}
\centering
\small
\setlength{\tabcolsep}{4pt}
\begin{tabular}{@{}lccccccccc|c@{}}
\toprule
\multicolumn{11}{c}{\textbf{Finger ($n{=}9$)}} \\
\midrule
Subject & S1 & S2 & S3 & S4 & S5 & S6 & S7 & S8 & S9 & \textbf{Mean} \\
$r_{\text{zero-shot}}$ & $0.252$ & $0.041$ & $0.045$ & $-0.017$ & $0.068$ & $0.154$ & $0.007$ & $0.025$ & $0.257$ & $\mathbf{0.092}$ \\
\bottomrule
\end{tabular}

\vspace{4pt}
\setlength{\tabcolsep}{2pt}
\resizebox{\textwidth}{!}{%
\begin{tabular}{@{}lcccccccccccccccc|c@{}}
\toprule
\multicolumn{18}{c}{\textbf{Audio ($n{=}16$)}} \\
\midrule
Subject & S1 & S2 & S3 & S4 & S5 & S6 & S7 & S8 & S9 & S10 & S11 & S12 & S13 & S14 & S15 & S16 & \textbf{Mean} \\
$r_{\text{zero-shot}}$ & $0.095$ & $-0.053$ & $0.246$ & $0.016$ & $0.172$ & $0.017$ & $0.206$ & $0.304$ & $-0.021$ & $0.117$ & $0.027$ & $0.193$ & $0.010$ & $0.010$ & $0.030$ & $0.424$ & $\mathbf{0.112}$ \\
\bottomrule
\end{tabular}%
}
\end{table}

\paragraph{Panel (f): Per-subject paired comparison at full data} For each subject, we plot a single dot at $(r_\text{persub}, r_\text{LOO-FT})$, both evaluated at $f{=}1.0$ on the held-out test split, on a single axis pooling both datasets. $r$ is the per-subject Pearson correlation, mean-over-fingers on the finger task. The dashed line is the identity $y=x$ and the shaded band $|\Delta r|{<}\delta$ ($\delta{=}0.05$) is treated as a tie. Three statistics are reported in the panel inset, all pooled across the 25 subjects: (i) Spearman rank correlation $\rho$ between $r_\text{persub}$ and $r_\text{LOO-FT}$ tests whether the two methods rank the same subjects in the same order; (ii) two-sided paired Wilcoxon signed-rank $p$ on $\Delta r$ tests whether their distributions are shifted; (iii) the win/tie/loss tallies count subjects with $\Delta r{>}\delta$ / $|\Delta r|{<}\delta$ / $\Delta r{<}{-}\delta$ respectively, providing a non-parametric per-subject readout. Subjects whose $|\Delta r|{\geq}\delta$ are individually labeled so that the outliers are inspectable.

\begin{table}[ht]
\caption{CORTEG architecture variants.}
\label{tab:arch}
\centering
\small
\begin{tabular}{@{}lcccccc@{}}
\toprule
\textbf{Variant} & \textbf{Depth} $L$ & \textbf{Embed} $D$ & \textbf{Heads} $H$ & \textbf{MLP ratio} & \textbf{Patch} $p_\text{lo}$ & \textbf{Params} \\
\midrule
Small & 8 & 512 & 8 & 4.0 & 16 & $\sim$25.6M \\
Base & 12 & 768 & 12 & 4.0 & 16 & $\sim$85.6M \\
Large & 24 & 1024 & 16 & 4.0 & 16 & $\sim$303.0M \\
\bottomrule
\end{tabular}
\end{table}

\begin{table}[ht]
\caption{Merge point sweep across model scales on the Stanford finger task. $K{=}\,$\texttt{hi\_inject\_last\_n} is the number of transformer layers after which the high-frequency stream is merged into the low-frequency stream; $K{=}L$ means merge at the input (all blocks see merged tokens). All entries are HPC runs at seed 42 with the extended HPC training (200 epochs, effective batch size 512 via batch 128, accumulation 4, patience 190); reported quantity is the val-best (final saved checkpoint) cross-subject mean Pearson $r$. The bolded best-of-$K$ per scale identifies the merge point used by Fig.~\ref{fig:decoding} (b) for all three scales (Small $K{=}1$, Base $K{=}2$, Large $K{=}6$).}
\label{tab:merge}
\centering
\small
\begin{tabular}{@{}cc@{\hspace{18pt}}cc@{\hspace{18pt}}cc@{}}
\toprule
\multicolumn{2}{c}{\textbf{Small} ($L{=}8$)} & \multicolumn{2}{c}{\textbf{Base} ($L{=}12$)} & \multicolumn{2}{c}{\textbf{Large} ($L{=}24$)} \\
$K$ & Mean $r$ & $K$ & Mean $r$ & $K$ & Mean $r$ \\
\midrule
1  & $\mathbf{0.577}$ & 2  & $\mathbf{0.583}$ & 3  & $0.576$ \\
2  & $0.576$          & 3  & $0.576$          & 6  & $\mathbf{0.577}$ \\
3  & $0.569$          & 5  & $0.568$          & 12 & $0.551$ \\
4  & $0.561$          & 6  & $0.558$          & 21 & $0.530$ \\
5  & $0.538$          & 8  & $0.525$          &    &                 \\
6  & $0.511$          & 9  & $0.517$          &    &                 \\
7  & $0.527$          & 11 & $0.470$          &    &                 \\
\bottomrule
\end{tabular}
\end{table}

     % A.4   app:decoding_protocols
%-----------------------------------------------------------------------------
\subsection{Experimental Protocols for Figure~\ref{fig:neuroscience}}
\label{app:manifold_protocols}
%-----------------------------------------------------------------------------

This subsection introduces how each panel of Figure~\ref{fig:neuroscience} was generated.

\paragraph{Panels (a, b): Per-electrode importance on a template cortex} The reliability-weighted importance score for each electrode of each subject is computed by a leave-one-channel-out sensitivity analysis: the trained pooled CORTEG-S model is evaluated on the held-out test split with one channel discarded at a time, and the per-channel drop in subject-mean Pearson $r$ is recorded as the raw importance. To make scores comparable across subjects with different baseline decoding strengths, we report the within-subject \emph{rank} of each channel ($\in[0,1]$, $1$=most important) multiplied by that subject's baseline test correlation $r_\mathrm{baseline}$, so a top-rank channel from a strong subject ($r_\mathrm{baseline}{\approx}0.7$) contributes more than a top-rank channel from a near-chance subject ($r_\mathrm{baseline}{\approx}0.1$). The color scale is shared across both datasets and clipped to a single $v_{\mathrm{max}}$ equal to the maximum baseline correlation seen across both. Stanford electrodes are drawn at the original coordinates provided with the dataset, which also serve as the model input, on the fsaverage template without warping them to its surface. Contacts that appear to lie off the cortex (most visibly for finger subject S8) are therefore an artifact of this template projection.

\paragraph{Panels (c, d): 3D PCA of token representations} For a representative finger-task subject (subject \texttt{bp}), we evaluate the trained model on the held-out test split with two configurations sharing identical architecture and training schedule but differing only in initialization: \emph{Pretrained} loads the pretrained small checkpoint into the backbone before fine-tuning; \emph{Random init} skips the load and begins from the standard Kaiming default. Both configurations use full fine-tuning with the KNNSoftFourier adapter (matched-protocol fairness control). For every test window, we extract the mean-pooled token representation $z\in\R^{D}$ ($D{=}512$ for the Small backbone) immediately before the regression head, yielding $T$ representation vectors. We center $z$ (subtracting the train-time mean) and fit \texttt{PCA(n\_components=3)} from \texttt{sklearn} on $z$, giving each timepoint a 3D coordinate. Points are colored by the dominant active finger: a window is 'active' if any of the 5 finger flexion targets exceeds an activity threshold ($z$-score $>1.0$); among active windows, the color is the index of the maximum-amplitude finger. Inactive (rest) windows are drawn first as faint grey, so colored active points sit on top. The participation ratio $\mathrm{PR}{=}(\sum\lambda_i)^2/\sum\lambda_i^2$, computed from the eigenvalues of the centered $z$ covariance, is annotated in the upper-left of each panel as a single scalar summary of effective dimensionality. Both panels share the same default matplotlib 3D viewing angle (azimuth $-60^\circ$, elevation $30^\circ$) and the same data preprocessing.

\paragraph{Panel (e): Cumulative explained variance} For each of the two configurations from panels (c, d), we fit \texttt{PCA(n\_components=30)} on the centered $z$ and plot the cumulative sum of the per-component explained-variance ratios. The dashed vertical line at $K{=}3$ matches the dimensionality of panels (c, d) and the legend annotates each curve with the cumulative variance at $K{=}3$. We use the same per-subject train-time normalization to make the two curves directly comparable.

\paragraph{Panel (f): Representational velocity around movement onset} Movement onsets are detected on the ground-truth finger trajectory: a frame $t$ is an onset for finger $f$ if $y_{t-1,f}{<}\tau$, $y_{t,f}{\geq}\tau$ ($\tau{=}1.0$ in $z$-units), and no other onset for the same finger has been registered in the prior $K{=}12$ frames (evaluation window @ 25 Hz). Across all 5 fingers, this yields $185$ onsets on the held-out test split for subject \texttt{bp}, of which $184$ have a full $[-K,+K]$ window and enter the average. For each configuration (pretrained, random) we project $z$ onto the top-3 PC subspace fitted on the centered $z$ from that configuration, compute the per-frame representational velocity $v_t = \|z^{(3)}_{t+1} - z^{(3)}_t\|_2$, and average $v_t$ across onsets in a window $[-K, +K]$ frames around each onset. The plotted line is the cross-onset mean, and the shaded band is $\pm 1$ standard error of the mean. The vertical dashed line at $t{=}0$ marks the onset frame; a steeper rise just before $t{=}0$ (as seen for the pretrained curve) reflects an earlier representational transition coincident with the upcoming movement.

     % A.5   app:manifold_protocols
%-----------------------------------------------------------------------------
\subsection{Per-Subject Electrode Implantations on MNI fsaverage}
\label{app:implantations}
%-----------------------------------------------------------------------------

\begin{figure}[ht]
\centering
\includegraphics[width=0.95\textwidth]{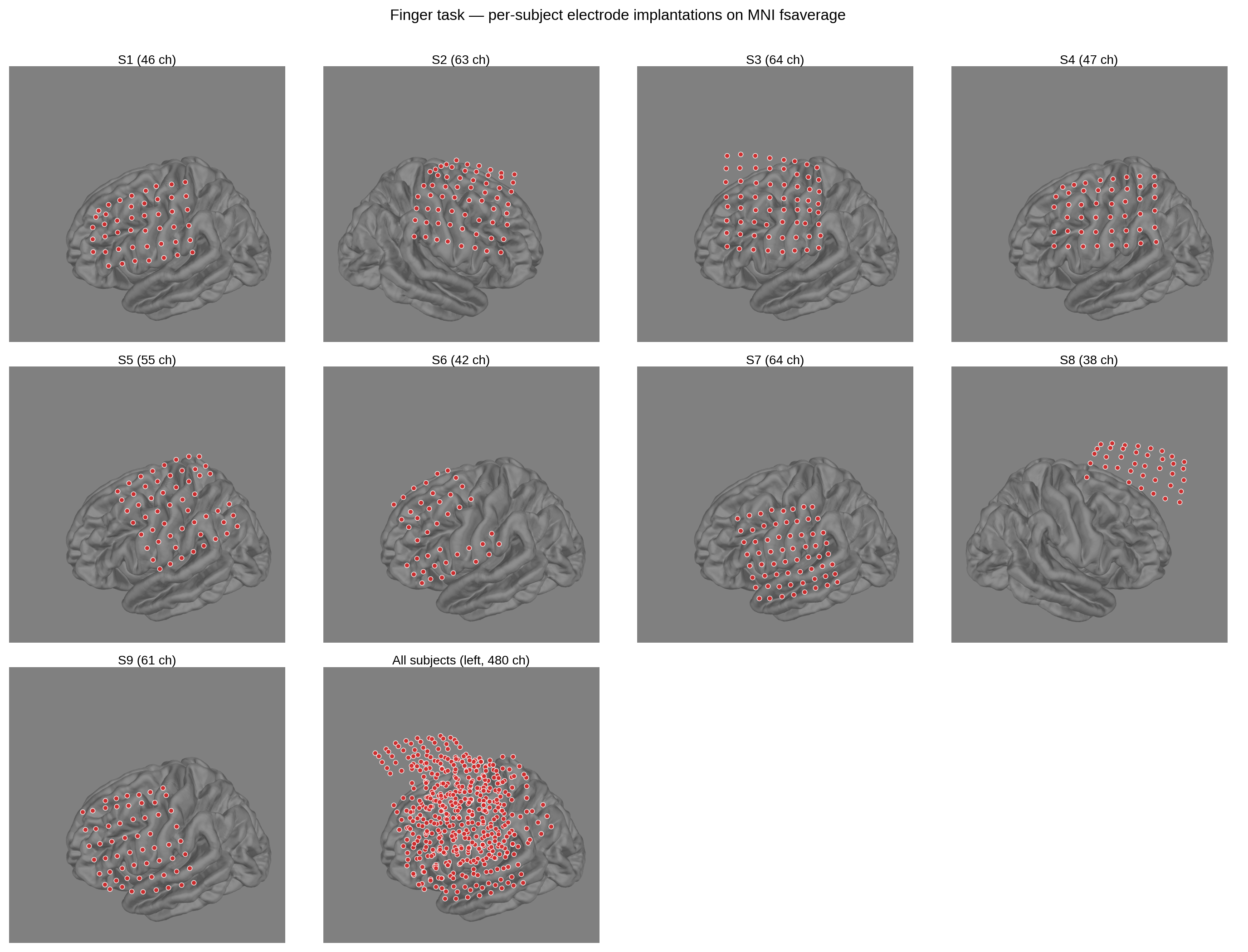}
\caption{Finger task: per-subject electrode implantations on MNI fsaverage (left- or right-lateral view chosen per subject by majority hemisphere). Panel S$i$ shows subject $i$'s grid; the final panel pools all 480 channels.}
\label{fig:implant_finger}
\end{figure}

\begin{figure}[ht]
\centering
\includegraphics[width=0.95\textwidth]{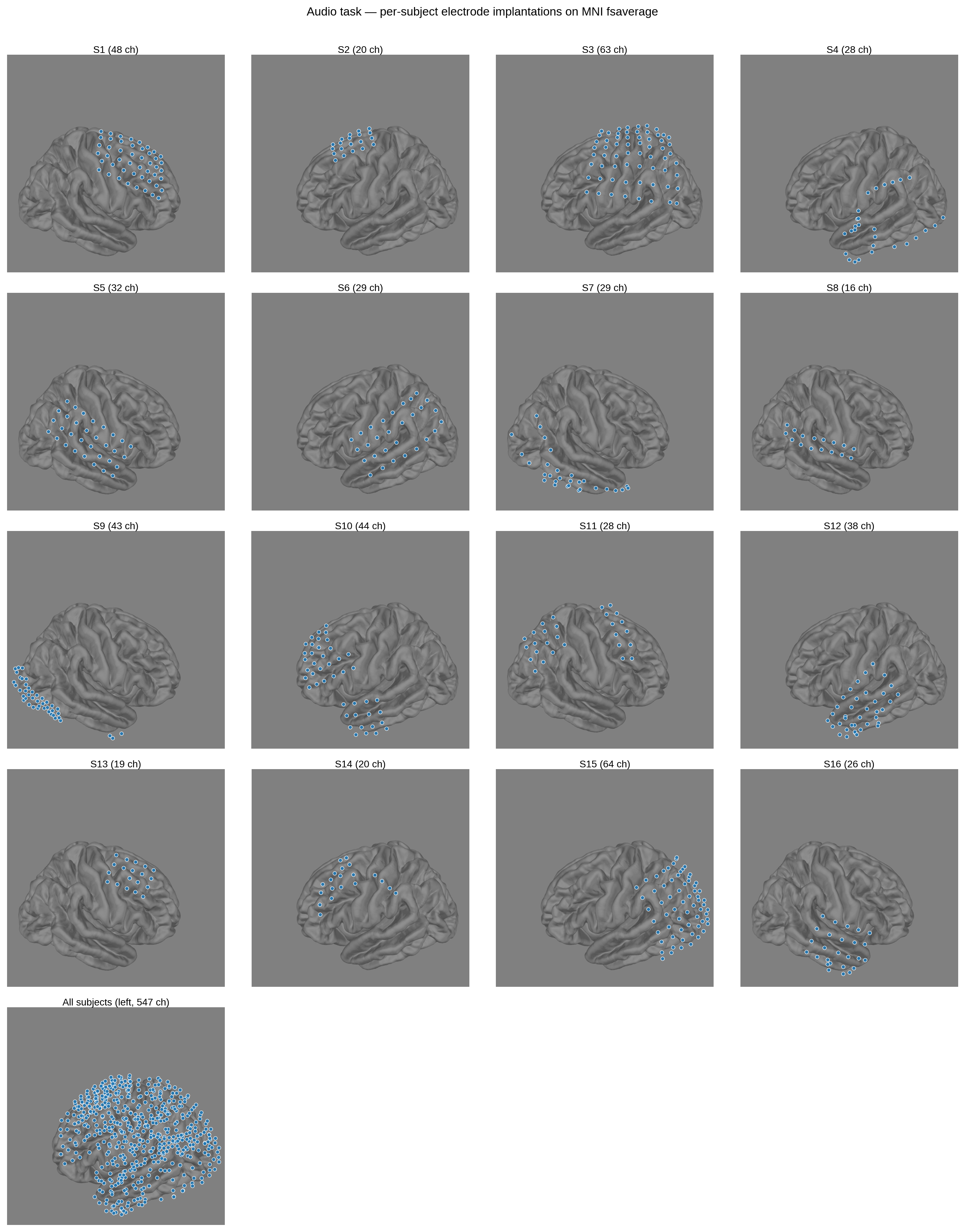}
\caption{Audio task: per-subject electrode implantations on MNI fsaverage. Panel S$i$ shows subject $i$'s electrodes; the final panel pools all 547 channels.}
\label{fig:implant_audio}
\end{figure}

Electrode positions for every subject in both datasets, projected onto the fsaverage MNI brain surface. Subjects are anonymized as $\mathrm{S1}$--$\mathrm{S}n$; per-panel titles show the channel count. The final panel of each figure pools all subjects' channels on the predominant hemisphere. The two datasets differ markedly in coverage: the finger task uses dense subdural grids over sensorimotor cortex (Fig.~\ref{fig:implant_finger}), while the audio task uses smaller, more variable strips over superior-temporal and Sylvian regions (Fig.~\ref{fig:implant_audio}).

\begin{figure}[ht]
\centering
\includegraphics[width=0.95\textwidth]{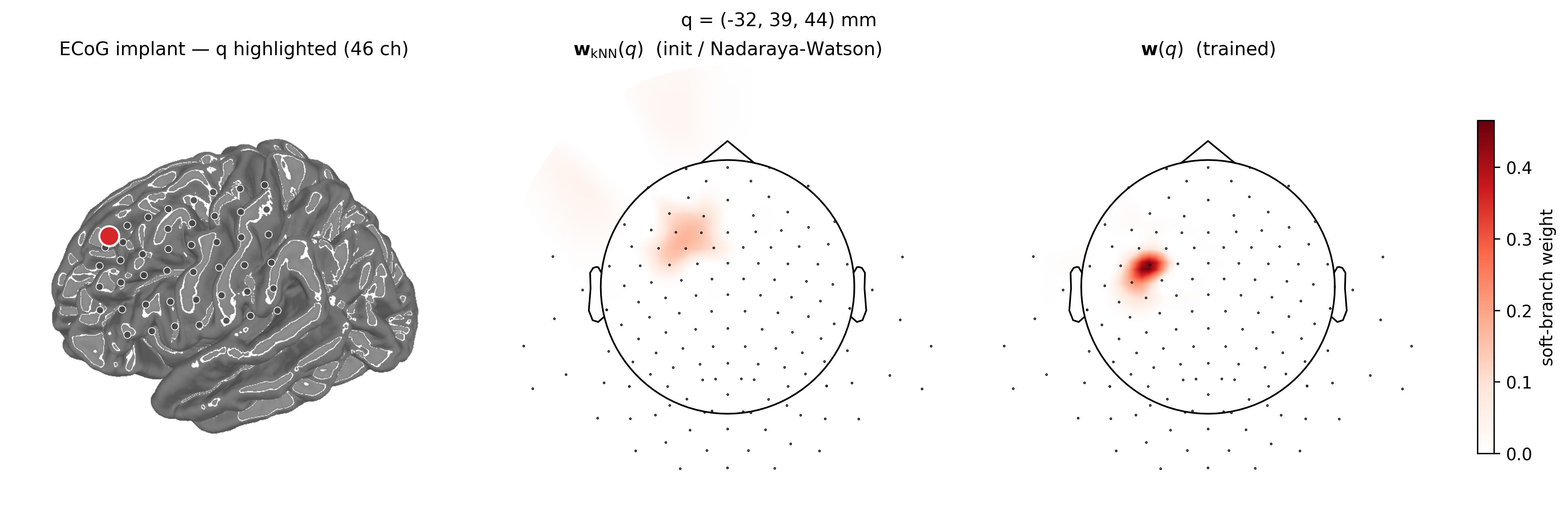}
\caption{Soft-branch attention $w(\q)$ for a representative ECoG electrode (subject \texttt{bp}, MNI $(-32, +39, +44)$~mm; left anterior frontal). \textbf{Left:} the subject's full ECoG implantation on \emph{fsaverage}, with the highlighted electrode in red. \textbf{Center:} attention at \emph{initialization}, equal to the $k$-NN Gaussian kernel over EEG positions ($k{=}8$, $\sigma{=}$ median nearest-neighbour distance) --- the Nadaraya--Watson estimator. \textbf{Right:} attention after training. Initialization spreads mass diffusely over $\sim$8 nearby EEG channels (top-3 weights $0.16/0.15/0.15$), while training concentrates it onto a single dominant channel (top-3 weights $0.47/0.22/0.09$).}
\label{fig:adapter_attention}
\end{figure}

\begin{figure}[ht]
\centering
\includegraphics[width=0.95\textwidth]{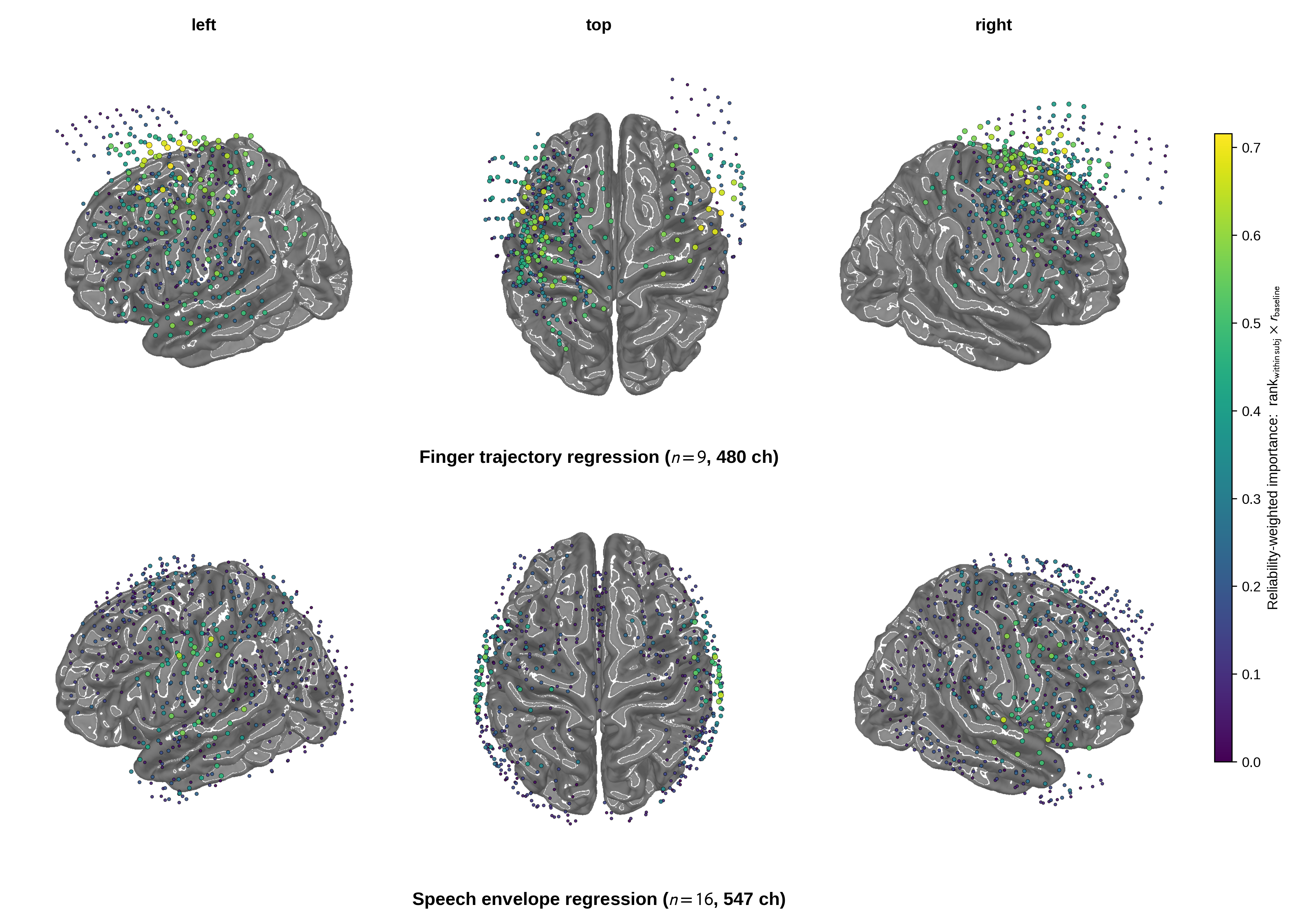}
\caption{Per-channel reliability-weighted importance $s_c = \mathrm{rank}_{\text{within\,subj}}(\Delta r_c) \times r_{\text{baseline}}^{(s)}$, pooled across subjects on fsaverage MNI (left lateral / dorsal / right lateral). Marker color and size both encode $s_c$ (yellow/large = high importance from a reliable subject; dark/small = low rank or unreliable subject). Top: finger trajectory ($n{=}9$, 480 channels). Bottom: audio envelope ($n{=}16$, 547 channels). The composite score makes anatomical clustering visible: left sensorimotor cortex on the finger task, bilateral superior-temporal / Sylvian regions on the audio task.}
\label{fig:channel_importance}
\end{figure}

          % A.6   app:implantations
%-----------------------------------------------------------------------------
\subsection{Alternative Projection Methods for the Neural Manifold}
\label{app:projections}
%-----------------------------------------------------------------------------

The main-text Figure~\ref{fig:neuroscience}~(c, d) uses unsupervised PCA on the mean-pooled token representation $z$. We additionally tried two supervised/structured alternatives common in motor-neuroscience analyses: Linear Discriminant Analysis (LDA)~\cite{hastie2009elements}, fit on active timepoints with the dominant active finger as label and projected onto the top three between-class discriminant axes (supervised by the displayed label, so the apparent compactness should not be interpreted as unsupervised cluster structure); and Demixed PCA (dPCA)~\cite{kobak2016demixed}, where latent representations $z$ were aligned to movement onset ($t\in[-12,12]$ frames) and grouped by finger, with at least $12$ trials per finger. dPCA separated condition-, time-, and interaction-related components, and the full latent trajectory was projected onto the top three condition-related components.

\begin{figure}[ht]
\centering
\includegraphics[width=\textwidth]{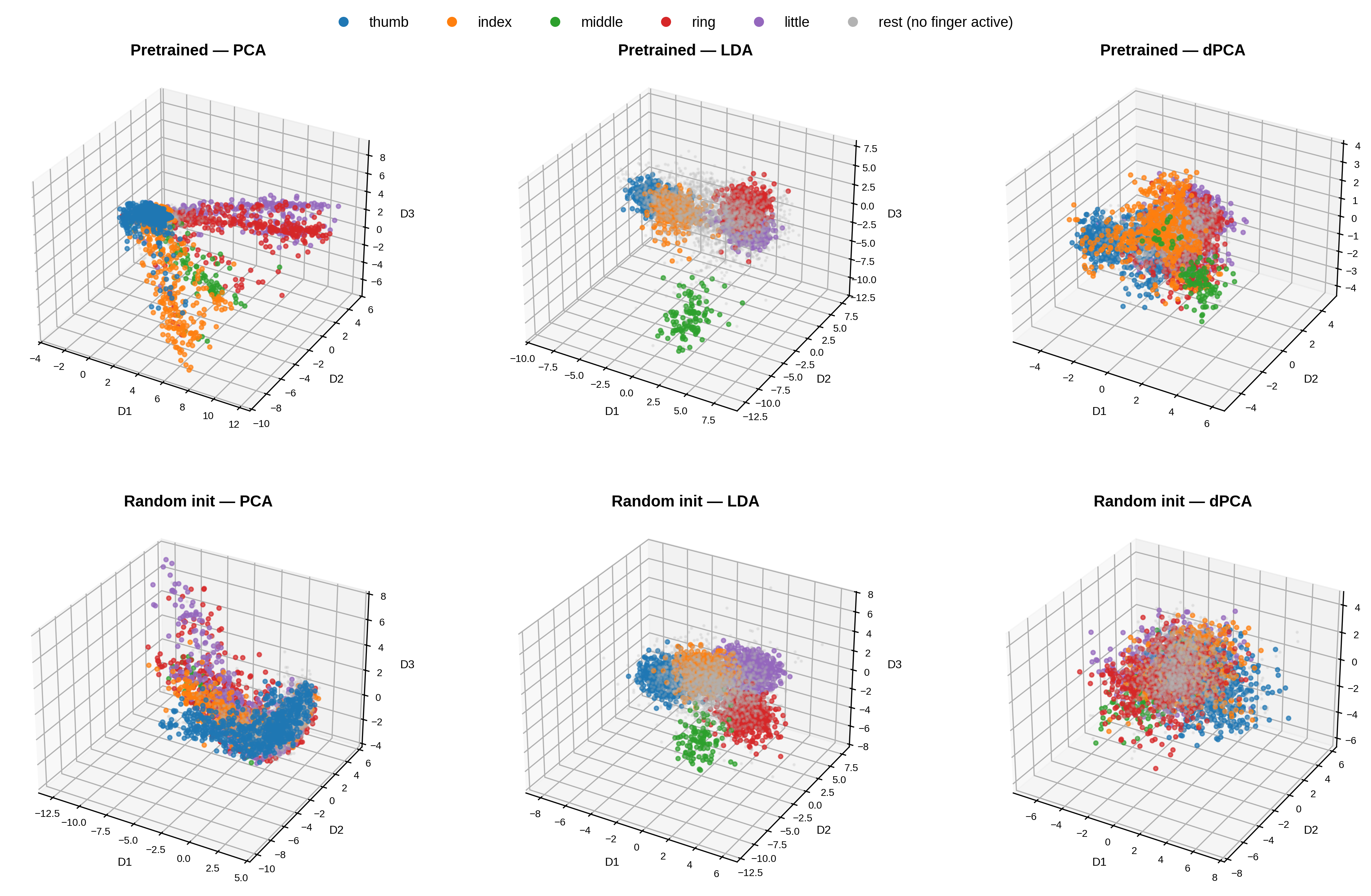}
\caption{Comparison of three projection methods for the neural manifold on a representative finger-task subject (full fine-tune). Top row: pretrained CORTEG. Bottom row: random-init CORTEG. Columns: PCA (unsupervised; used in Figure~\ref{fig:neuroscience}), LDA (supervised by the displayed finger label), and dPCA (condition-related axes after onset-aligned trial averaging). LDA gives the visually strongest finger separation but is supervised by the same label that is displayed; PCA and dPCA are unsupervised at projection time. All three methods agree qualitatively that the pretrained representation supports cleaner finger-conditional structure than the random-init baseline. Grey points are rest timepoints (no finger above the activity threshold).}
\label{fig:manifold_projections}
\end{figure}

All three projections agree on the qualitative picture (Figure~\ref{fig:manifold_projections}): the pretrained backbone produces a finger-discriminable representation while random init does not. We use the unsupervised PCA panels in the main text because they (i) match the V0-paper convention, (ii) do not bias toward the displayed label, and (iii) co-vary directly with the unsupervised participation-ratio and variance-concentration statistics reported in Figure~\ref{fig:neuroscience} (e). The LDA panels are included here to show the upper-bound separability, and the dPCA panels show that the pretrained-vs-random gap survives even when condition-related axes are isolated from time-related and noise components.

   % A.7   app:projections
%-----------------------------------------------------------------------------
\subsection{Ablation Details: Backbone Model Ablation and Input-Stream Controls}
\label{app:ablation}
%-----------------------------------------------------------------------------

Unless otherwise noted, all rows in Table~\ref{tab:ablation} follow the pooled-training protocol described in \S\ref{sec:setup}. The three exceptions are marked in the table caption. The random-initialization control ($^{\star}$) uses full fine-tuning rather than LoRA, since LoRA alone cannot meaningfully adapt a randomly initialized backbone. The remaining two controls ($^{\P}$ and $^{\mathsection}$) use different strategies, as detailed below.

\paragraph{REVE backbone ($^{\P}$)}
REVE~\cite{el2026reve} is a $69$M-parameter EEG foundation model pretrained on approximately $25{,}000$ subjects. To test whether CORTEG depends on a particular pretrained backbone, we retain the overall CORTEG recipe, using dual-stream tokenization, mid-layer fusion, LoRA adaptation of the trailing blocks, and geometry conditioning, replacing only ST-EEGFormer with REVE. Since REVE incorporates electrode coordinates through its own positional pathway, CORTEG's KNNSoftFourier adapter is replaced by REVE's native coordinate mechanism.

The resulting model reaches $r{=}0.529$ on finger decoding and $r{=}0.293$ on audio decoding, corresponding to $95\%$ and $86\%$, respectively, of the ST-EEGFormer-based CORTEG scores and remaining well above the raw high-gamma ridge baseline ($0.336$ and $0.175$, Table~\ref{tab:main}). This indicates that the effectiveness of the proposed method is not specific to a single pretrained backbone, although the remaining performance gap suggests that backbone choice still matters. The task-dependent stream dissociation in Table~\ref{tab:ablation} also reproduces with REVE: high-gamma outperforms low-frequency input on finger decoding by $+0.162$ ($9/9$ subjects, $p{=}0.004$), whereas low-frequency input outperforms high-gamma on audio decoding by $+0.098$ ($13/16$ subjects, $p{=}0.003$).

\paragraph{Matched single-broadband control ($^{\mathsection}$)}
The single-stream ablations in Table~\ref{tab:ablation} compare the dual-stream model with models that receive only one frequency range, thereby testing the contribution of each band rather than the benefit of explicitly decomposing the signal into two streams. To isolate the latter, we additionally compare against a matched single-stream broadband model operating directly on the native $256$ Hz recording, which retains the high-gamma frequencies. The broadband model uses the same token width and trainable-parameter count as its dual-stream counterpart.

For finger decoding, this control was evaluated under a matched single-seed protocol in which the corresponding dual-stream model achieved $r{=}0.545$, close to the reported $0.554$. Accordingly, the $\Delta{=}{-}0.011$ reported in Table~\ref{tab:ablation} is computed relative to this matched run rather than to the $0.554$ value in the main ablation row. The difference is not significant ($5/9$ subjects favor dual-stream, $p{=}0.57$).

       % A.8   app:ablation
%-----------------------------------------------------------------------------
\subsection{Fusion Strategies (Empirical)}
\label{app:fusion}
%-----------------------------------------------------------------------------

The proposed CORTEG model fuses the low-frequency and high-gamma streams via a parameter-free mean-pooling operation applied to the merged tokens after the dual-stream backbone. We tested a range of more sophisticated alternatives before settling on this choice. Table~\ref{tab:fusion} summarizes the full sweep. None of the per-merge-point variants in this sweep improved over the baseline. The one exception---a layer-wise gated injection that gates the high-gamma stream into every block rather than merging once---is analyzed in App.~\ref{app:gate} (Table~\ref{tab:layerwise}): it is neutral at Small scale on Stanford (the setting of this table), gains significantly at Stanford-Base and Ghent-Base/Large, and regresses slightly at Ghent-Small. Each single-merge variant is briefly described below.

\paragraph{Routing mechanisms} \emph{SPVAE latent router.} Two small VAE encoders map $z_\text{lo}$ and $z_\text{hi}$ into shared and private latent components $(z_\text{shared}, z_\text{lo}^\text{priv}, z_\text{hi}^\text{priv})$ with KL regularization against a standard-normal prior; a per-channel precision router then mixes the two streams with weights derived from a product-of-experts on the diagonal Gaussian posteriors. \emph{HPR v1--v3 (heteroscedastic precision routing).} Three variants of a no-VAE alternative: each stream gets a lightweight prediction head that emits both the target and a per-channel log-variance $\sigma^2$, calibrated via heteroscedastic Gaussian NLL; the router uses Bayesian-optimal mixing weights $\alpha_\text{ch} = \text{prec}_\text{hi}/(\text{prec}_\text{lo}+\text{prec}_\text{hi})$ with $\text{prec}=1/\sigma^2$. v1--v3 differ in calibration regularizers (representation-agreement loss, variance scaling, etc.). \emph{Learned gating $\alpha(z)\in[0,1]$.} A scalar gate per token, predicted by a small MLP from the merged representation, that linearly interpolates: $z_\text{merged}=\alpha\,z_\text{hi}+(1-\alpha)\,z_\text{lo}$. \emph{Cross-attention merge.} Replaces the mean-pool with a cross-attention block in which one stream forms queries and the other supplies keys and values, allowing tokens of one stream to attend over the other's representation before pooling.

\paragraph{Auxiliary fusion losses} \emph{CID (sufficiency + complementarity).} Two auxiliary heads attached to the per-stream pooled representations enforce a mutual-information chain decomposition $I(y;z_\text{lo},z_\text{hi})=I(y;z_\text{lo})+I(y;z_\text{hi}\mid z_\text{lo})$ via two MSE terms: a \emph{sufficiency} loss $\|\hat{y}_\text{lo}(z_\text{lo})-y\|^2$ pushing the lo-stream to be predictive of the target on its own, and a \emph{complementarity} loss $\|\hat{y}_\text{hi}(z_\text{hi})-(y-\hat{y}_\text{lo})\|^2$ (with stop-gradient on the residual) pushing the hi-stream to predict only what lo cannot. The main regression loss is unchanged; CID is added as an auxiliary objective.

\begin{table}[ht]
\caption{Fusion-strategy variants on the Stanford finger task, pooled training, $\text{seed}{=}42$. Mean Pearson $r$ across the $n{=}9$ subjects; $\Delta$ is the change vs.\ the baseline. All variants share the remaining CORTEG-S configuration. Entries are HPC-protocol runs at the default merge point $K{=}4$. The corresponding 5090-protocol number is reported in Table~\ref{tab:main} ($r{=}0.554$). The HPC-vs-local gap is consistent with the reproducibility note in App.~\ref{app:decoding_protocols}.}
\label{tab:fusion}
\centering
\small
\setlength{\tabcolsep}{6pt}
\begin{tabular}{@{}lcc@{}}
\toprule
\textbf{Variant} & \textbf{Mean $r$} & \textbf{$\Delta$ (pp)} \\
\midrule
\textbf{Baseline} (mean-pool late fusion) & $\mathbf{0.573}$ & --- \\
\midrule
\textit{Layer-wise gating (multi-seed)} & & \\
\quad Layer-wise gated injection (Small; $+1.8$\,pp at Base, Table~\ref{tab:layerwise}) & $0.575$ & $+0.2$ \\
\midrule
\textit{Routing mechanisms (per-token gating across the two streams)} & & \\
\quad SPVAE latent router                          & $0.573$ & $\phantom{-}0.0$ \\
\quad HPR v1--v3 (precision routing)               & $\leq 0.573$ & $\leq 0.0$ \\
\quad Learned gating $\alpha(z) \in [0,1]$         & $0.535$ & $-3.8$ \\
\quad Cross-attention merge                        & $0.404$ & $-16.9$ \\
\midrule
\textit{Auxiliary fusion losses} & & \\
\quad CID (sufficiency + complementarity loss)     & $0.548$ & $-2.5$ \\
\bottomrule
\end{tabular}
\end{table}

\paragraph{Layer-wise gate design} The layer-wise gated injection of App.~\ref{app:gate} is robust to its own hyperparameters. Table~\ref{tab:gate_design} sweeps the gate activation, bottleneck width, and whether the per-block gate is shared across blocks. All variants land within ${\approx}0.006$ of one another and of the default gate, and rank within seed noise. The largest deviation is sharing a single gate across all blocks.

\begin{table}[ht]
\caption{Layer-wise gate design ablation (Stanford finger, pooled, Small backbone, $n_\text{seed}{=}4$). Mean per-subject Pearson $r\pm$ cross-seed SD. The default gate uses a tanh activation, bottleneck $16$, and an independent gate per block.}
\label{tab:gate_design}
\centering
\small
\begin{tabular}{@{}lc@{}}
\toprule
\textbf{Design variant} & \textbf{Mean $r$} \\
\midrule
Linear gate (no activation)     & $0.571 \pm 0.006$ \\
Sigmoid activation (vs.\ tanh)  & $0.570 \pm 0.003$ \\
Bottleneck $64$                 & $0.571 \pm 0.009$ \\
Bottleneck $8$                  & $0.571 \pm 0.006$ \\
Gate shared across blocks       & $0.569 \pm 0.009$ \\
\bottomrule
\end{tabular}
\end{table}

\paragraph{LoRA depth and rank}
Table~\ref{tab:lora_sens} sweeps the number of trailing transformer blocks adapted by LoRA and the rank $r$. The default configuration (last 4 blocks, $r{=}4$) is at or near the maximum across both axes; deeper LoRA modestly underperforms, and increasing the rank from $4$ to $8$ does not help. We did not observe any configuration in this sweep that exceeded the default by more than the seed-noise margin.

\begin{table}[ht]
\caption{LoRA depth and rank sweep on the finger task pooled (small backbone, KNNSoftFourier adapter). Default configuration in \textbf{bold}.}
\label{tab:lora_sens}
\centering
\small
\begin{tabular}{@{}lcc@{}}
\toprule
\textbf{Configuration} & \textbf{Mean $r$} & \textbf{$\Delta$ vs.\ default} \\
\midrule
\textbf{Last 4 blocks, $r{=}4$ (default)} & $\mathbf{0.573}$ & --- \\
Last 6 blocks, $r{=}4$                    & $0.565$ & $-0.008$ \\
Last 8 blocks, $r{=}4$                    & $0.563$ & $-0.010$ \\
Last 4 blocks, $r{=}8$                    & $0.554$ & $-0.019$ \\
\bottomrule
\end{tabular}
\end{table}

      % A.9   app:fusion
%-----------------------------------------------------------------------------
\subsection{Layer-wise Gated Fusion}
\label{app:gate}
%-----------------------------------------------------------------------------

The default CORTEG architecture merges the low-frequency (LFS) and high-gamma (HGA) streams once, using a parameter-free mean operation at block $L{-}K$ (\S\ref{sec:method:problem}). To avoid the hyperparameter sweep, we also propose a \emph{layer-wise gated injection} strategy. Rather than merging the streams at a single depth, a small gating MLP (${\approx}17$ K parameters) reads pooled LFS and HGA representations and produces a scalar gate $g_\ell$ for each transformer block. After block $\ell$, the HGA representation is injected into the low-frequency stream through the residual update $z \leftarrow z + g_\ell z_{\mathrm{hi}}$. The last layer of the gating MLP is zero-initialized, so with the default tanh activation, every $g_\ell$ is $0$ at the start, and optimization begins from the low-frequency-only model.

Table~\ref{tab:layerwise} compares layer-wise gating with mean-pool fusion on both datasets and at all three backbone scales under the multi-seed HPC protocol (seeds $\{0,1,2,3,42\}$). Statistical testing follows the paper's subject-level paired Wilcoxon procedure ($n{=}9$ finger and $n{=}16$ audio subjects). We first average each subject's score across seeds, so seeds are not treated as independent observations. The results show that the benefit of gating is conditional, not universal. Gating significantly improves performance for Stanford-Base and for Ghent-Base and Ghent-Large. It is indistinguishable from mean-pool at Stanford-Small, shows a positive but non-significant trend at Stanford-Large, and produces a small but significant decrease at Ghent-Small.

The layer-wise gated variant offers two practical advantages over the default mean-pool fusion. First, it removes the need to select a single merge depth \(K\), allowing high-gamma information to be incorporated adaptively across the transformer layers. Second, it improves performance in several settings, with the clearest gains observed for the Base and Large models on the Ghent dataset, while remaining competitive elsewhere. However, these improvements are not consistent across all dataset--scale combinations and are modest in most cases. We therefore retain parameter-free mean-pooling as the default CORTEG design for its simplicity and robustness, and report layer-wise gated fusion as an alternative lightweight variant that can provide additional gains in some settings at a small parameter cost.

\begin{figure}[t]
\centering
\includegraphics[width=\textwidth]{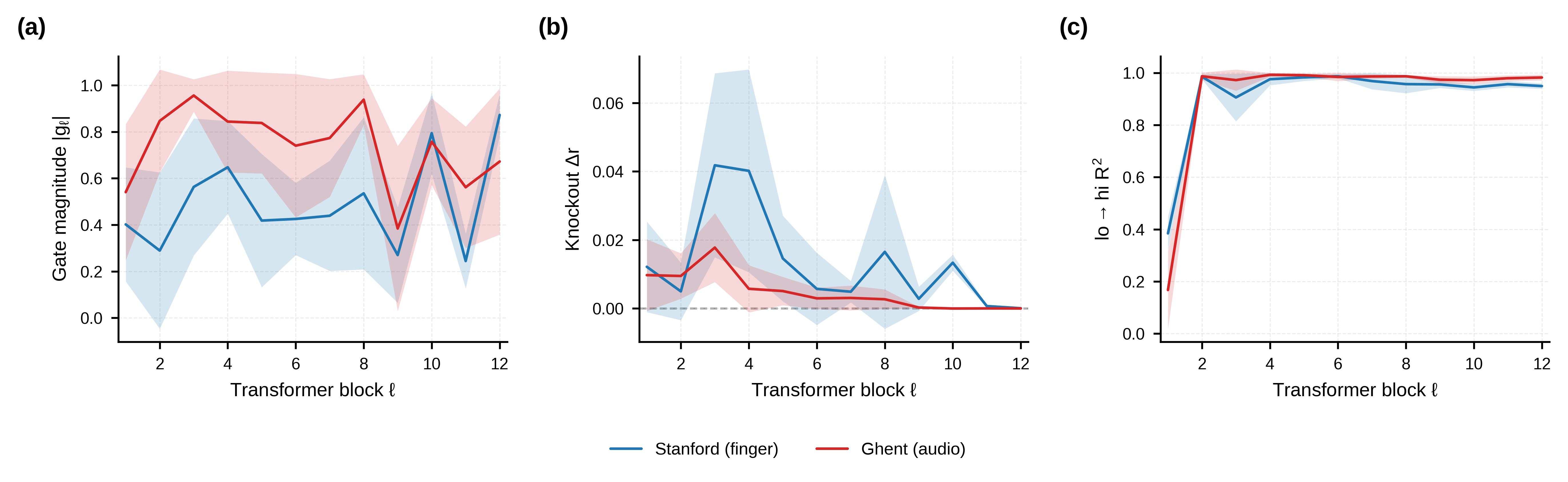}
\caption{Layer-wise gate analysis (Base backbone; mean $\pm$ SD over 5 seeds $\{0,1,2,3,42\}$).
Per-block analysis of the trained gated model on Stanford (finger) and Ghent (audio).
\textbf{(a)} Mean gate magnitude $|g_\ell|$, showing that HGA information is injected at multiple depths.
\textbf{(b)} Knockout $\Delta r$, defined as the decrease in test correlation when the HGA injection at block $\ell$ is removed. The largest contributions occur in the early-to-middle blocks, peaking at block~3 on both datasets and diminishing at later depths.
\textbf{(c)} lo$\rightarrow$hi linear $R^2$ from a ridge probe. The LFS representation becomes increasingly predictive of HGA after successive gated injections, reaching $R^2{\approx}0.98$ by mid-depth, indicating progressive integration of HGA information into the LFS stream.}
\label{fig:gate}
\end{figure}

We further analyze the learned per-layer gating mechanism in Figure~\ref{fig:gate} using the Base backbone (mean $\pm$ SD over $5$ seeds). A leave-one-block-out ablation of the HGA injection shows that its decoding contribution is concentrated in the early-to-middle layers, peaking at block~3 on both datasets and becoming small at later layers, consistently across seeds. This suggests that the gated formulation can exploit HGA information at depths where a single predefined merge cannot. Panel~(c) further shows that repeated gated injection makes the LFS representation increasingly predictive of HGA, reaching lo$\rightarrow$hi ridge $R^2{\approx}0.98$ by mid-depth. This suggests that the fusion mechanism progressively transfers HGA information into the LFS stream.

\begin{table}[ht]
\caption{Layer-wise gated fusion versus the default mean-pool fusion under pooled training with the multi-seed HPC protocol. Values are mean subject-level Pearson $r$ ($\pm$ cross-seed SD; $n_{\mathrm{seed}}{=}5$ for Small/Base and $3$ for Large). $\Delta$ denotes gate $-$ mean-pool. We assess statistical significance using the paper's subject-level paired Wilcoxon test ($n{=}9$ finger / $16$ audio subjects), after averaging each subject's score across seeds. Reported $p$-values are uncorrected. \textbf{Bold} marks comparisons that remain significant after correction across the six cells.}
\label{tab:layerwise}

\centering
\small
\setlength{\tabcolsep}{6pt}
\begin{tabular}{@{}llccc@{}}
\toprule
\textbf{Dataset} & \textbf{Scale} & \textbf{Mean-pool} &
\textbf{Layer-wise gate} & \textbf{$\Delta$ ($p$)} \\
\midrule
\multirow{3}{*}{Stanford (finger)}
 & Small & $0.572 {\pm} 0.006$ & $0.575 {\pm} 0.006$ & $+0.002\ (0.73)$ \\
 & Base  & $0.573 {\pm} 0.009$ & $\mathbf{0.591 {\pm} 0.007}$ &
 $\mathbf{+0.018\ (0.008)}$ \\
 & Large & $0.569 {\pm} 0.007$ & $0.579 {\pm} 0.016$ & $+0.010\ (0.039)$ \\
\midrule
\multirow{3}{*}{Ghent (audio)}
 & Small & $\mathbf{0.337 {\pm} 0.008}$ & $0.322 {\pm} 0.007$ &
 $\mathbf{-0.015}\ (0.013)$ \\
 & Base  & $0.237 {\pm} 0.118$ & $\mathbf{0.281 {\pm} 0.057}$ &
 $\mathbf{+0.044\ ({<}0.001)}$ \\
 & Large & $0.194 {\pm} 0.130$ & $\mathbf{0.323 {\pm} 0.007}$ &
 $\mathbf{+0.129\ ({<}0.001)}$ \\
\bottomrule
\end{tabular}
\end{table}

         % A.10  app:gate
%-----------------------------------------------------------------------------
\subsection{Intracranial Foundation Model Comparison}
\label{app:fm}
%-----------------------------------------------------------------------------

We expand the iEEG-FM comparison of Table~\ref{tab:main} using BrainBERT~\cite{wang2023brainbert}, PopT~\cite{chau2025population}, and Brant~\cite{zhang2023brant}. For BrainBERT and PopT, released pretrained weights are evaluated with a common \texttt{extract}$\rightarrow$head pipeline under frozen linear probing and last-$N$ block fine-tuning, each tested with pooled, per-subject, and LOO training, and under a trained temporal head, tested with pooled and per-subject training. Stanford is evaluated at its native $1$\,kHz to preserve the high-gamma content expected by BrainBERT, whereas Ghent is evaluated at $128$\,Hz (Nyquist $64$\,Hz). On Ghent, the FMs therefore do not see the $64$--$124$\,Hz part of the $60$--$124$\,Hz band used by CORTEG's high-gamma stream and by the raw high-gamma ridge baseline; this does not account for the gap, since CORTEG restricted to its low-frequency stream still reaches $r{=}0.273$ on audio (Table~\ref{tab:ablation}). PopT additionally receives electrode coordinates.

BrainBERT follows its released frozen recipe, which mean-pools the center $10$ spectrogram frames (PopT's per-electrode input tokens are pooled the same way). In our causal $1$\,s windows, their centroid lies ${\approx}500$\,ms before the target at the window's right edge; on a model-free proxy (ridge on raw high-gamma over the same windows, three subjects), center pooling lowers $r$ by ${\approx}34\%$ relative to pooling the full window. The BrainBERT and PopT heads use their own training budgets rather than CORTEG's: at most $60$ epochs with patience $20$ for the probe and for last-$N$ fine-tuning (last $2$ blocks, learning rate $10^{-4}$), and $100$ epochs with patience $20$ for the temporal head, a two-layer bidirectional LSTM over non-overlapping chunks of $64$ consecutive windows (and hence non-causal within a chunk); all three use batch size $256$. The FMs receive the raw signal and apply their own preprocessing, while the splits, windows, and train-only z-scoring of targets are shared with CORTEG. Table~\ref{tab:fm} summarizes the best adaptation for each model, and Table~\ref{tab:fm_full} gives the corresponding results.

\begin{table}[ht]
\caption{Intracranial-FM adaptation on continuous ECoG regression.
Mean Pearson $r\pm$cross-subject SD ($n{=}9$ finger, $16$ audio). Each cell reports the best pooled, per-subject, or LOO result for that adaptation. $^{\P}$ denotes three-seed experiments, averaged within subject before aggregation; all others use seed $42$. \textbf{Bold} marks the best adaptation within each model and task. In $11/14$ cells, the cross-subject SD exceeds the mean, and no configuration reaches $r{=}0.07$, compared with CORTEG at $0.554/0.339$ and raw high-gamma ridge at $0.336/0.175$.}
\label{tab:fm}
\centering
\footnotesize
\setlength{\tabcolsep}{5pt}
\begin{tabular}{@{}llcc@{}}
\toprule
\textbf{Model} & \textbf{Adaptation} & \textbf{Finger ($r$, $n{=}9$)} & \textbf{Audio ($r$, $n{=}16$)} \\
\midrule
\multirow{3}{*}{BrainBERT~\cite{wang2023brainbert}}
 & frozen linear probe    & $0.044 {\pm} 0.051$          & $0.052 {\pm} 0.067^{\P}$          \\
 & last-$N$ fine-tune     & $0.047 {\pm} 0.038$          & $0.050 {\pm} 0.087^{\P}$          \\
 & temporal head (BiLSTM) & $\mathbf{0.053} {\pm} 0.056$ & $\mathbf{0.057} {\pm} 0.066$      \\
\midrule
\multirow{3}{*}{PopT~\cite{chau2025population}}
 & frozen linear probe    & $0.050 {\pm} 0.045$          & $\mathbf{0.039} {\pm} 0.064$      \\
 & last-$N$ fine-tune     & $0.041 {\pm} 0.046^{\P}$     & $0.021 {\pm} 0.043^{\P}$          \\
 & temporal head (BiLSTM) & $\mathbf{0.063} {\pm} 0.046$ & $0.028 {\pm} 0.051$               \\
\midrule
Brant~\cite{zhang2023brant}
 & frozen probe ($6$\,s context) & $\mathbf{0.028} {\pm} 0.031$ & $\mathbf{0.022} {\pm} 0.035$ \\
\bottomrule
\end{tabular}
\end{table}

\begin{table}[ht]
\caption{Regression results for BrainBERT and PopT (mean Pearson $r$). Multi-seed cells marked $^{\P}$ in Table~\ref{tab:fm} are averaged within subject before aggregation. Stanford uses native $1$\,kHz input and Ghent $128$\,Hz. CORTEG pooled reference: $0.554$ finger / $0.339$ audio.}
\label{tab:fm_full}
\centering
\small
\setlength{\tabcolsep}{5pt}
\begin{tabular}{@{}llcccc@{}}
\toprule
\textbf{Dataset} & \textbf{Regime} & \textbf{BrainBERT probe} & \textbf{PopT probe} & \textbf{BrainBERT ft} & \textbf{PopT ft} \\
\midrule
\multirow{3}{*}{Stanford ($1$\,kHz)}
 & pooled      & $0.030$ & $0.040$ & $0.035$ & $0.041$ \\
 & per-subject & $0.039$ & $0.049$ & $0.033$ & $0.037$ \\
 & LOO         & $0.044$ & $0.050$ & $0.047$ & $0.036$ \\
\midrule
\multirow{3}{*}{Ghent ($128$\,Hz)}
 & pooled      & $0.031$ & $0.031$ & $0.028$ & $0.015$ \\
 & per-subject & $0.052$ & $0.038$ & $0.050$ & $0.021$ \\
 & LOO         & $0.050$ & $0.039$ & $0.043$ & $0.013$ \\
\bottomrule
\end{tabular}
\end{table}

We further evaluate several adaptation strategies for the iEEG FMs. Per-electrode readouts and full-backbone fine-tuning improve some individual subjects, but the mean performance remains low. Adding a two-layer BiLSTM to capture temporal context yields similarly limited gains, reaching $0.053/0.057$ for BrainBERT and $0.063/0.028$ for PopT on finger/audio decoding. Joint fine-tuning of the foundation model and temporal decoder does not further improve performance. Overall, although individual adaptation choices can improve specific configurations, the tested iEEG FMs remain weak on these continuous regression tasks.

Brant~\cite{zhang2023brant} requires a fixed $6$\,s, $250$\,Hz input patch, so we evaluate it separately using its native preprocessing and context. For each target, the preceding $6$\,s of ECoG is resampled to $250$\,Hz, converted to Brant's eight-band log-PSD representation, and passed through the released temporal and spatial encoders. Embeddings extracted on a causal $0.5$\,s grid yield $r{=}0.028$ on finger and $0.022$ on audio; a finer $0.1$\,s grid gives a similar finger result ($0.025$). Although the Ghent recordings cannot provide Brant's $70$--$128$\,Hz bands, the model is otherwise evaluated with its native context and preprocessing, suggesting that its poor transfer reflects a mismatch between the learned representation and these continuous decoding tasks rather than insufficient input context.

A representation-level analysis supports this interpretation. On Stanford, frozen BrainBERT and PopT embeddings ($r{\approx}0.05$) perform well below both raw high-gamma ($r{=}0.336$) and low-frequency ($r{=}0.181$) ridge baselines. Moreover, neither embedding linearly reconstructs the high-gamma power of its own input window, despite retaining global broadband-power and subject information. This pattern suggests that the pretrained representations do not preserve the fast, electrode-local high-gamma variations most relevant to continuous finger decoding.

Finally, on BrainTreebank (\S\ref{sec:results:btb}), the same pretrained models perform substantially better on intracranial classification tasks. This suggests that their weak performance above is specific to continuous regression rather than a general failure of the pretrained representations.

\begin{table}[ht]
\caption{BrainTreebank results under the leakage-free protocol.
Mean AUROC ($\pm$ cross-subject SD) across $10$ subjects using four-fold chronological forward chaining with a $7$\,s gap. Task~A is sentence-initial vs.\ mid-sentence word classification. Task~B is word vs.\ non-word classification. The final column reports cross-seed SD of the cohort mean where multiple seeds are available.}
\label{tab:btb}
\centering
\footnotesize
\setlength{\tabcolsep}{5pt}
\newcommand{\btbgrp}[1]{\multicolumn{4}{@{}l}{\textcolor{gray}{\textit{#1}}} \\}
\begin{tabular}{@{}lccc@{}}
\toprule
\textbf{Arm}
& \textbf{Task A}
& \textbf{Task B}
& \textbf{Seed SD} (A\,/\,B) \\
\midrule

\btbgrp{Raw spectral features (linear probe)}
Raw high-gamma, per-channel
& $0.5846{\pm}0.0296$
& $0.8155{\pm}0.0623$
& --- \\
Raw band-power, per-channel
& $0.5796{\pm}0.0344$
& $0.7978{\pm}0.0728$
& --- \\
Raw high-gamma, channel-mean
& $0.5127{\pm}0.0230$
& $0.5217{\pm}0.0485$
& --- \\

\midrule
\btbgrp{CORTEG (pooled, per-subject LoRA)}
\textbf{CORTEG} (layer-wise gate)
& $0.6376{\pm}0.0783$
& $0.7492{\pm}0.1345$
& $0.0135$\,/\,$0.0146$ \\
CORTEG (mean-pool fusion)
& $0.6159{\pm}0.0721$
& $0.7525{\pm}0.1326$
& $0.0318$\,/\,$0.0036$ \\
CORTEG, random-init backbone$^{\star}$
& $0.5376{\pm}0.0536$
& $0.5880{\pm}0.1025$
& $0.0031$\,/\,$0.0184$ \\

\midrule
\btbgrp{Trained from scratch}
HiLoFuseNet
& $0.5330{\pm}0.0220$
& $0.7208{\pm}0.0886$
& $0.0217$\,/\,$0.0018$ \\
CNN-LSTM
& $0.5071{\pm}0.0142$
& $0.5706{\pm}0.0469$
& $0.0055$\,/\,$0.0032$ \\
LSTM
& $0.5073{\pm}0.0136$
& $0.5618{\pm}0.0429$
& $0.0036$\,/\,$0.0006$ \\

\midrule
\btbgrp{PopT~\cite{chau2025population}}
PopT, LoRA
& $0.6003{\pm}0.0845$
& $0.7790{\pm}0.0991$
& --- \\
PopT, full fine-tune
& $0.6001{\pm}0.0726$
& $0.6979{\pm}0.1228$
& --- \\
PopT, head-only
& $0.5594{\pm}0.0442$
& $0.6714{\pm}0.0915$
& --- \\
PopT, frozen probe
& $0.5697{\pm}0.0358$
& $0.6698{\pm}0.0814$
& --- \\

\midrule
\btbgrp{BrainBERT~\cite{wang2023brainbert}}
BrainBERT, single-elec. max
& $0.6015{\pm}0.0450$
& $0.6980{\pm}0.0941$
& --- \\
BrainBERT, population mean-pool
& $0.5294{\pm}0.0280$
& $0.5788{\pm}0.0664$
& --- \\
BrainBERT, full fine-tune
& $0.5480{\pm}0.0343$
& $0.5682{\pm}0.0775$
& --- \\
BrainBERT, head-only
& $0.5446{\pm}0.0258$
& $0.5483{\pm}0.0563$
& --- \\
BrainBERT, LoRA
& $0.5463{\pm}0.0243$
& $0.5412{\pm}0.0534$
& --- \\
BrainBERT, single-elec.\ mean
& $0.5085{\pm}0.0063$
& $0.5236{\pm}0.0166$
& --- \\

\midrule
\btbgrp{Brant~\cite{zhang2023brant} ($505$M parameters)}
Brant, single-elec.\ max
& $0.5865{\pm}0.0415$
& $0.5775{\pm}0.0366$
& --- \\
Brant, head-only
& $0.5245{\pm}0.0168$
& $0.5423{\pm}0.0183$
& --- \\
Brant, LoRA
& $0.5174{\pm}0.0147$
& $0.5322{\pm}0.0268$
& --- \\
Brant, full fine-tune
& $0.5310{\pm}0.0377$
& $0.5283{\pm}0.0314$
& --- \\
Brant, single-elec.\ mean
& $0.5216{\pm}0.0114$
& $0.5261{\pm}0.0107$
& --- \\
Brant, population mean-pool
& $0.5208{\pm}0.0158$
& $0.5233{\pm}0.0120$
& --- \\

\bottomrule
\end{tabular}

\vspace{2pt}
{\footnotesize
$^{\star}$Random initialization is a pretraining control with the same architecture, folds, training recipe, seeds, and fusion as CORTEG, but without loading the pretrained checkpoint.}
\end{table}
 % A.11  app:fm

%%%%%%%%%%%%%%%%%%%%%%%%%%%%%%%%%%%%%%%%%%%%%%%%%%%%%%%%%%%%

\clearpage % flush all pending floats from the appendix

\end{document}